\documentclass[lettersize,journal]{IEEEtran}
\usepackage{amsmath,amsfonts}
\usepackage{algorithmic}
\usepackage{algorithm}
\usepackage{array}
\usepackage{textcomp}
\usepackage{stfloats}
\usepackage{url}
\usepackage{verbatim}
\usepackage{graphicx}
\usepackage{cite}
\usepackage{color}

\usepackage[table]{xcolor}
\usepackage{multirow}
\usepackage{booktabs} 
\usepackage{graphicx} 
\usepackage{float} 
\usepackage{subfigure} 

\begin{document}

\title{CL-BioGAN: Biologically-Inspired  Cross-Domain Continual Learning for Hyperspectral Anomaly Detection}

\author{
        Jianing~Wang$^{*}$, \IEEEmembership{Member, IEEE,}
        Zheng Hua$^{*}$,
        Wan Zhang,
        Shengjia~Hao,
        Yuqiong~Yao,
        and Maoguo~Gong, \IEEEmembership{Fellow, IEEE}
\thanks{
Manuscript received X X, 2024; accepted X X, 2025. This work was supported in part by the National Natural Science Foundation of China under Grant 61801353.

Jianing Wang and Wan Zhang is with the Key Laboratory of Collaborative Intelligence Systems, Ministry of Education, School of Computer Science and Technology, Xidian University, Xi’an 710071, China.

Zheng Hua, Shengjia Hao and Yuqiong Yao are with the Key Laboratory of Intelligent Perception and Image Understanding of Ministry of Education of China, School of Artificial Intelligence, Xidian University, Xi’an 710071, China. (Corresponding author: Jianing Wang, e-mail: circuitwang@163.com; Wan Zhang, e-mail: zhangwan2021@163.com)

Maoguo Gong is with the Key Laboratory of Collaborative Intelligence Systems, Ministry of Education, Xidian University, Xi'an 710071, China. (e-mail: Gong@ieee.org)
}
\thanks{$^{*}$Jianing Wang and Zheng Hua contributed equally to this work. (e-mail: circuitwang@163.com, HuaZheng79@163.com)}
}

\markboth{Journal of \LaTeX\ Class Files,~Vol.~14, No.~8, July~2024}%
{Shell \MakeLowercase{\textit{et al.}}: A Sample Article Using IEEEtran.cls for IEEE Journals}


\maketitle

\begin{abstract}

Memory stability and learning flexibility in continual learning (CL) is a core challenge for cross-scene Hyperspectral Anomaly Detection (HAD) task. Biological neural networks can actively forget history knowledge that conflicts with the learning of new experiences by regulating learning-triggered synaptic expansion and synaptic convergence. Inspired by this phenomenon, we propose a novel Biologically-Inspired Continual Learning Generative Adversarial Network (CL-BioGAN) for augmenting continuous distribution fitting ability for cross-domain HAD task, where Continual Learning Bio-inspired Loss (CL-Bio Loss) and self-attention Generative Adversarial Network (BioGAN) are incorporated to realize forgetting history knowledge as well as involving replay strategy in the proposed BioGAN. Specifically, a novel Bio-Inspired Loss  composed with an Active Forgetting Loss (AF Loss) and a CL loss is designed to realize parameters releasing and enhancing between new task and history tasks from a Bayesian perspective. Meanwhile, BioGAN loss with $\textbf L_{2}$-Norm enhances self-attention (SA) to further balance the stability and flexibility for better fitting background distribution for open scenario HAD (OHAD) tasks. Experiment results underscore that the proposed CL-BioGAN can achieve more robust and satisfying accuracy for cross-domain HAD with fewer parameters and computation cost. This dual contribution not only elevates CL performance but also offers new insights into neural adaptation mechanisms in OHAD task.
\end{abstract}

\begin{IEEEkeywords}
Continual learning, Hyperspectral anomaly detection, Active forgetting, GAN.
\end{IEEEkeywords}

\section{Introduction}
\begin{figure}[htb] 
\centering 
\includegraphics[scale=0.45]{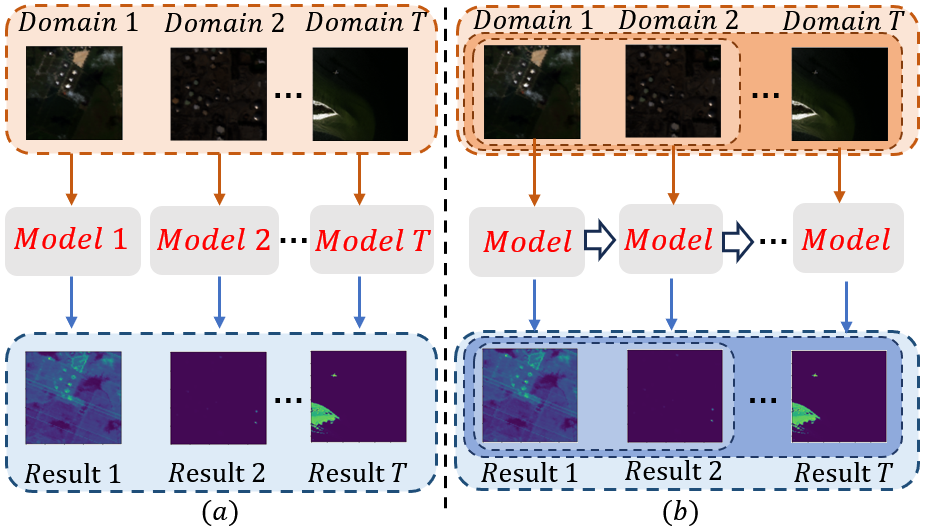} 
\caption{Comparison of traditional DL training model and the CL method. (a) represents the traditional DL method, which obtains OHAD results by a set of independent well-trained parameters. (b) represents the proposed CL method. The parameters of the model are continuously updated with the arrived tasks, but the updated parameters will not forget the previous learned knowledge.}
\label{CL} 
\end{figure}

\IEEEPARstart{H}{yperspectral} image (HSIs) contains hundreds or even thousands of contiguous narrow spectral bands, which makes possibilities for precisely distinguishing different materials \cite{77}. As one of important research fields of hyperspectral information processing, the main aim of OHAD task is to find pixels that are significantly different from the background in terms of spectral signatures without any prior knowledge of target \cite{-1}. Recent years, various traditional deep learning (DL)-based algorithms have been proposed for OHAD tasks \cite{0}. 
However, most existing DL-based methods mainly excel at acquiring knowledge through generalized learning behavior based on solving specific scene task from a distinct training phase \cite{78}. As shown in Fig.\ref{CL} (a), traditional deep learning (DL)-based algorithms can only excel at specific task or current scenario because of the specific constructed parameters of network is incapable of dealing with new tasks or scenarios, which is liable to result in catastrophic forgetting phenomenon. Whereas the network parameters of CL can be automatically updated through customized loss functions or automatically updated exemplar set. After the training for $t$ different scenarios, the learned parameters of the model contains the ability of OHAD for all the previous tasks in open scenario circumstance, the procedure is briefly illustrated in Fig.\ref{CL} (b).

Continual Learning (CL) \cite{1} (also known as Lifelong Learning \cite{2} or Incremental Learning \cite{87}) as a representative method for alleviating catastrophic forgetting phenomenon caused by varied conditions and scenarios \cite{4}, is gradually attracting more and more attention in recent years \cite{88}. Aim at learning knowledge and feature properties of current task while still maintaining the knowledge information from previous training tasks, current methodologies in CL can be conceptually grouped into three main categories: 1). Adding regularization terms with reference to the old model (regularization-based approach); 2). Approximating and recovering the old data distributions (replay-based approach); and 3). Constructing task-specific parameters with a properly-designed architecture (architecture-based approach) \cite{-7}, which have presented proficiencies in addressing numerous CL challenges. 
However, the effectiveness of strategies varied significantly across different experimental settings, such as task types and similarities, input sizes, and training sample quantities \cite{79}. Therefore, a significant gap exists between research advancements and practical applications especially for OHAD tasks.

\begin{figure*}[htb] 
\centering 
\includegraphics[height=8cm,width=15cm]{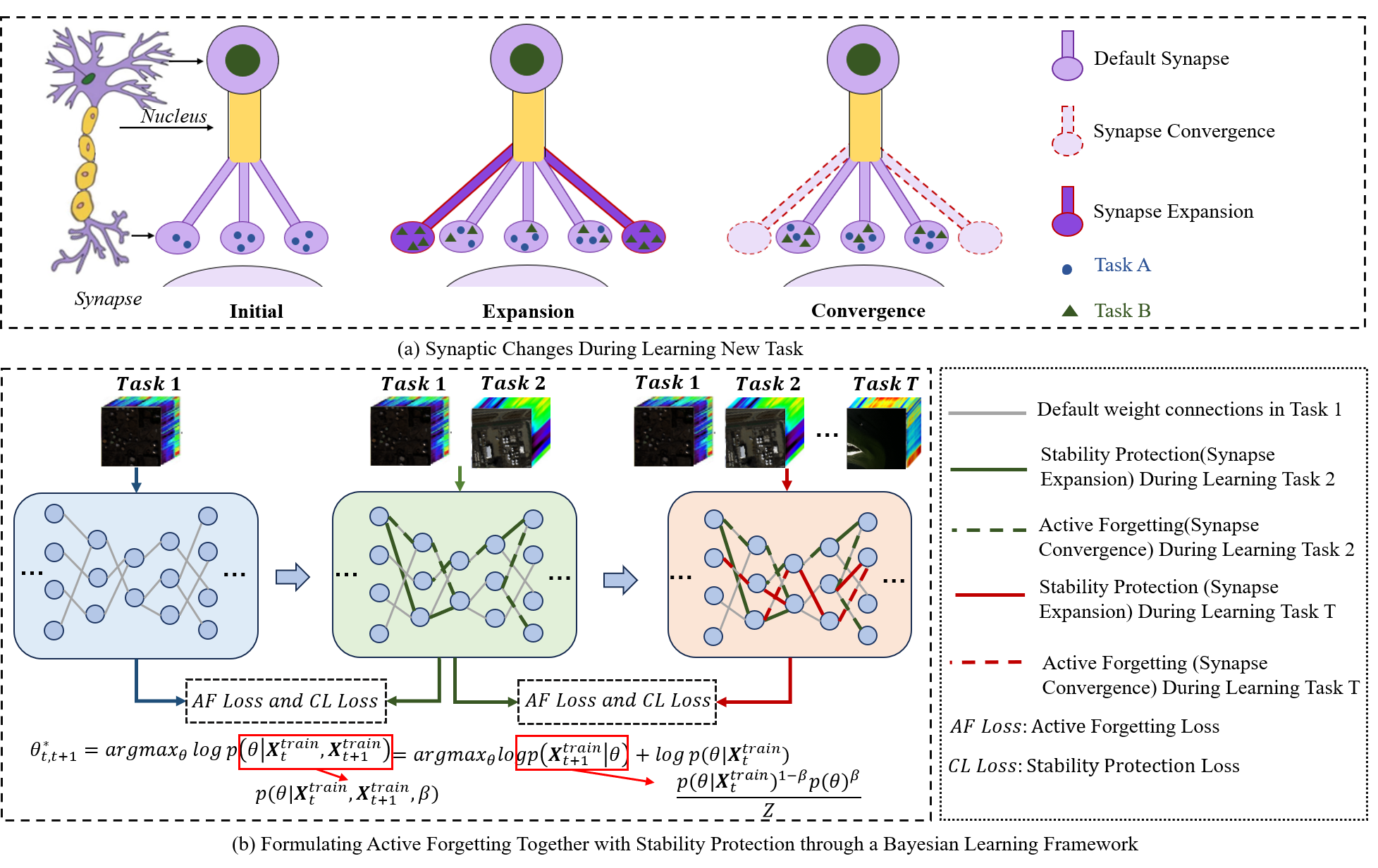} 
\caption{Principles of active forgetting and learning in our proposed neural network. (a) represents the Neural and Protrusion Function Diagram. 
(b) represents the Synaptic Changes Diagram During the Learning of a New Task.} 
\label{tuchu} 
\end{figure*}

Biological neural networks can effectively learn new knowledge based on memorizing history knowledge, even if they conflict with each other. This advantage of memory flexibility is achieved by actively forgetting history knowledge that interferes with the learning of new experiences \cite{wang2023incorporating}. Inspired by this biological properties, in order to alleviate catastrophic forgetting problem caused by traditional DL-based method, we explore CL from a biological learning perspective. As shown in Fig. \ref{tuchu} (a), the biological active forgetting properties can regulate the learning-triggered synaptic expansion and synaptic convergence \cite{6}. Specifically, biological synapses can expand with additional functional connections to learn new experiences with previously learned functional connections (synaptic expansion), and then prune excrescent connections by transferring useful knowledge to the previously learned synaptic connections (synaptic pruning). 
Driven by this phenomenon, the proposed CL-BioGAN can actively forget history knowledge which interferes with new learning tasks without significantly arousing catastrophic forgetting. We formulate this process with the framework of Bayesian learning to well synthesize and model biological synaptic plasticity by tracking the probability distribution of synaptic weights under dynamic sensory inputs \cite{7},\cite{8}.
Specifically, we employ Bayesian CL incorporate SA to actively forget the posterior distribution, absorbing all information from the history tasks with a forgetting factor to better learn each new task for facilitating CL performance, which fully utilizes bio-inspired loss and replay strategy to mitigate the issue of catastrophic forgetting. Besides, the potential spatial feature extraction capability of $\textbf L_{2}$-norm enhanced SA and the local feature extraction capability of CNN are designed through GAN structure to reconstruct the background pixels. 

The main innovations and contributions of the proposed CL-BioGAN are summarized as follows:
\begin{itemize}
\item A novel CL-BioGAN is proposed to elaborately integrating CL-Bio Loss and BioGAN structure, which can adaptively efficient releasing ineffective parameters as well as strengthening robust parameters through local-global Transformer-based SA  learning. CL-BioGAN can provide a dual stable and flexible way from Bio-inspired network structure and Bio-inspired Loss to achieve high accuracy with robust performance for cross-domain OHAD task.
\item The proposed BioGAN model is incorporated with a $\textbf L_{2}$ local-global SA discriminator structure, which can efficiently approximate the background distribution through alternating training with Bio-inspired Encoder-Decoder Generator to acquire local-global representative background spectral characteristics.
\item A novel Bio-Inspired Loss is designed to integrate BioGAN loss, AF loss and CL loss for balancing the memory stability and learning flexibility for OHAD. The BioGAN loss updates the parameters of network for current task and enhances the robustness of BioGAN. AF loss is formulated with the framework of Bayesian learning, and encourages the model to improve the probability of learning each new task through attenuating old memories by a designed forgetting rate. Meanwhile, the CL loss can automatically balance and maintain the stability of parameters in learning procedure.
\end{itemize}

The proposed CL-BioGAN can realize efficient generation and fitting ability to the varied background distribution, which presents satisfying detection accuracy and robust performance for OHAD, and offers a novel approach to the application of DL for open scenarios remote sensing area.

The rest of this article is organized as follows. The related work is introduced in Section II. The detail of the proposed CL-BioGAN framework is introduced in Section III. The related experiments are analyzed and presented in Section IV. Conclusion and future work are presented in Section V.

\section{Related Work }

{\bf{Continual learning.}} Modern machine learning often outperforms human intelligence when it comes to developing robust models from fixed data sets and fixed surroundings \cite{9}. Traditional DL techniques usually suffer from catastrophic forgetting phenomenon for forgetting what has been previously learned by overriding with new data \cite{14-2}.
Therefore, CL methods address this issue by striking a balance between the stability (ability to retain knowledge) and plasticity (ability to learn new concepts) of neural network models \cite{14-3}. For deep learning-based CL, similar to the learning of human, the learner typically encounters a non-stationary data stream through a series of learning episodes \cite{14-1}.

Numerous efforts have been devoted to preserving memory stability to mitigate catastrophic forgetting in artificial neural networks. The realization of existing CL methods can be roughly categorized into five mainstreams.
{\bf{1). Replay-based techniques:}} These approaches typically create a replay pool in which features or samples from earlier tasks are maintained to avoid catastrophic forgetting. Generative models is proposed to generate pseudo data for replay strategy \cite{11,12}. Cheng et al. \cite{13} select the Augmentation Stability Rehearsal (ASR) method by estimating the augmentation stability to select the most representative and discriminative samples. Constructive noises is introduced in each stage of constructive vision transformer (ViT) and an exponential moving average of the working modal is utilized to enforce consistency in predictions \cite{14}. An exemplar-based CL method is introduced by using contrastive learning to learn a task-agnostic and continuously improved feature expression \cite{90}.
Rehearsal-based methods aim at handling ‘stability-plasticity’ dilemma by maintaining an appropriate balance between learning new information and retaining previous knowledge \cite{16-1}.
{\bf{2). Regularization-based techniques:}} These techniques are mainly based on minimizing the catastrophic forgetting through utilizing the regularization loss term to minimize the model's parameter distance in various tasks.
A dual distillation training objectives is proposed to achieve the combination of two independent models of old classes and new classes \cite{17}. Class Similarity Knowledge Distillation (CSW-KD) method is proposed to perform semantic segmentation by extracting the knowledge of previous models of old classes which are similar to the new class\cite{18}. 
Three contrastive learning losses are designed to significantly improve the CL effect in the field of remote sensing images \cite{19}. In \cite{20}, the elastic weight-consolidated loss function is combined with a self-supervised network. Focusing on the CL model to retain old knowledge while ensuring the model's ability to learn new knowledge, Kim et al. \cite{21} proposed Auxiliary Network Continual Learning (ANCL). Regularization-based methods can effectively avoid the storage of old tasks, therefore it can avoid data leakage and the need for increased memory.
{\bf{3). Architecture-based methods:}} In order to reduce forgetting performance, these methods mainly involve new task-specific layers or structures for new tasks and activities. Parameter masking \cite{22} can learn binary masks on existing networks to obtain a suitable neural network for multiple tasks. Li et al. \cite{23} proposed a method to use architecture search for the best structure of each task. Meta-attention (MEAT) mechanism is designed to determine which image patches and parameters are needed to be isolated by optimizing the binary mask \cite{24}. In \cite{25}, the network structure are divided into task-sharing components and task-specific components, where task-specific components can be continuously expanded for new coming tasks. This type of method is adept at dealing with increasing domain shift tasks. In addition, a rule-based model ensemble \cite{86} is introduced by forming a distributed architecture to manage high-dimensional data, where an adjustable weight vector is emphasized by the role of correlation strategies in incremental learning. 
{\bf{4). Optimization-Based techniques:}} optimization-based CL algorithms can be achieved by explicitly designing and manipulating the optimization programs. A typical idea is to perform gradient projection. For instance, AdamNSCL \cite{92} instead projects candidate parameter updates into the current null space approximated by the uncentered feature covariance of the old tasks, while AdNS \cite{93} considers the shared part of the previous and the current null spaces. Another attractive idea is meta-learning or learning-to-learn for continual learning, which attempts to obtain a data-driven inductive bias for various scenarios, rather than designing it manually \cite{94}. ARI \cite{95} combines adversarial attacks with experience replay to obtain task-specific models, which are then fused together through meta-training. MARK \cite{96} maintains a set of shared weights that are incrementally updated with meta-learning and selectively masked to solve specific tasks. Besides, some other works refine the optimization process from a loss landscape perspective. Linear Connector \cite{97} adopts Adam-NSCL \cite{98} and feature distillation to obtain respective solutions of the old and new tasks connected by a linear path of low error, followed by linear averaging.
{\bf{5). Representation-Based techniques:}} have attempted to incorporate the advantages of self-supervised learning (SSL) and large-scale pre-training to improve the representations in initialization and in CL. Based on self-supervised learning strategy (basically with contrastive loss) for CL, MinRed \cite{99} further promotes the diversity of experience replay by decorrelating the stored old training samples. CaSSLe \cite{100} converts the self-supervised loss to a distillation strategy by mapping the current state of a representation to its previous state. Based on pre-training for downstream continual learning, representative strategies include selecting the most relevant prompts from a prompt pool (L2P \cite{101}), optimizing a weighted summation of the prompt pool with attention factors (CODAPrompt \cite{102}), using explicitly task-sharing and task-specific prompts (DualPrompt \cite{103}) or only task-specific prompts (SPrompts \cite{104}), progressive expansion of task-specific prompts (Progressive Prompts \cite{105}), etc.

{\bf{Hyperspectral anomaly detection.}} HAD task aims at detecting the anomalies from background samples through the significantly difference between continuous spectral signatures \cite{26}. Existing HAD algorithms can be roughly divided into three mainstreams: {\bf{1) Statistic-based HAD approaches}} mainly estimate the probability of anomalous samples primarily by calculating the relationship difference between the background and the anomalous distribution by calculating differences. 
Reed-Xiaoli (RX) \cite{26} is proposed to detect anomalies by computing the mean and variance of the entire image through comparing Mahalanobis distance between the test sample and the background mean vector. Many variations based on RX, such as kernel-based RX algorithm (KRX) \cite{26-1}, local RX (LRX) \cite{26-3} have been gradually proposed. 
Meanwhile, a tensor RX (TRX) algorithm based on Fractional Fourier Transform \cite{27} and the Recursive RX with Extended Multi-Attribute Profiles \cite{28} are designed to extract more image feature information.
In \cite{32}, an adaptive reference correlation graph embedding (ARGE) is proposed to effectively obtain low-dimensional features and improve computational efficiency. To overcome the limitations of distributional assumptions in HAD, recent years have seen increased interest in {\bf{2) Representation-based HAD approaches}}. The cooperative representation-based detector (CRD) \cite{37} assumes that background pixels can be minimized by enhancing the $\textbf L_{2}$-norm of the representation weight vector, approximately represented by a linear combination of its spatial neighbors. LRASR \cite{38} assumes that backgrounds present low-rank attributes and anomalies present sparse attributes. Chang et al. \cite{39} proposed an Effective Anomaly Space (EAS) to solve the problem of anomalies being sandwiched between the background and noise during the background suppression process in HAD. Total Variation (TV) representation model is introduced into HAD\cite{40}, such as the new enhanced total variation (ETV) of the endmember background dictionary (EBD) to represent the row vector of the coefficient matrix \cite{41}. Feedback Band Group and Variation Low-Rank Sparse Model (FBGVLRS-AD) is proposed by utilizing local spatial constraints within spectral features combined with total variation tensor-based HAD method \cite{42}.
{\bf{3) Deep learning-based HAD approaches}} can learn complex data distribution even under the condition of lack of prior information \cite{43}, therefore background estimation and reconstruction are key issues in the implementation of HAD. Based on the ability for hierarchical learning, abstraction, and high-level representation of  Autoencoder (AE), which provides an unsupervised manner for background reconstruction \cite{44}. The background-guided Deformable Convolutional AutoEncoder (DCAE) network is designed for HAD \cite{45}. In \cite{46}, a weighted adaptive loss combined with AE is designed to suppress the reconstruction of abnormal samples. PASSNet \cite{105} built a lightweight hybrid model, which ensembles the respective inductive bias from convolutional neural network (CNNs) and global receptive field from transformers with designed patch attention module (PAM) to extract spatial–spectral features from multiple spatial perspectives by integrating both CNNs and transformers blocks.
In terms of feature representation capabilities and adversarial training capabilities, Generative Adversarial Network (GAN) has been successfully used to estimate background distribution and spectral domain features to achieve high-confidence background sample reconstruction \cite{47}, \cite{wang2021dual}. Under the condition that the number of background samples is much larger than abnormal samples, the recognition error of abnormal pixels is higher than that of background pixels. Jiang et al. \cite{48} only use normal background samples to train GAN to estimate the background distribution to achieve the detection of abnormal samples. Based on the visual attention Transformer model, a clustering module is introduced to detect false background and abnormal samples in \cite{49}. In 
\cite{106}, a hyperspectral anomaly detection network based on variational background inference and generative adversarial framework (VBIGAN-AD) is established based on the relationship between data samples and latent samples through two sub-networks to capture the data distribution.

{\bf{Bayesian Continual Learning.}} CL needs to remember the old tasks and learn each new task effectively. Therefore, considering a neural network with parameter $\theta $ continually learns two independent task $t$ and task $t+1$ based on their training datasets ${\textbf X_t^{train}}$ and ${\textbf X_{t+1}^{train}}$. All the training dataset for each task is only available when learning the task, therefore the posterior distribution and the Bayesian CL can be represented as follow.
\paragraph{Bayesian Learning} After learning ${\textbf X_t^{train}}$, the posterior distribution:
\begin{equation}
\begin{split}
\label{eq02}
p(\theta | \textbf X_t^{train}) = \frac{{p(\textbf X_t^{train}|\theta )p(\theta )}}{{p(\textbf X_t^{train})}}
\end{split}
\end{equation}
incorporates the knowledge of task $t$. Then, we can get the predictive distribution for the test data of task $t$:
\begin{equation}
\begin{split}
\label{eq12}
p(\textbf X_t^{test}|\textbf X_t^{train}) = \int {p(\textbf X_t^{test}|\theta )p(\theta |\textbf X_t^{train})} d\theta
\end{split}
\end{equation}
where as the posterior $p(\theta |\textbf X_t^{train})$ is generally intractable (except very special cases), we must resort to approximation methods, such as the Laplace approximation or other approaches of approximate inference \cite{83}. Let's take Laplace approximation as an example. If $p(\theta |\textbf X_t^{train})$ is smooth and majorly peaked around the mode $\theta _t^* =  \arg {\max _\theta }\log p(\theta |\textbf X_t^{train})$, we can approximate it with a Gaussian distribution whose mean is $\theta _t^*$ and covariance is the inverse Hessian of the negative log posterior\cite{6}.
\paragraph{Bayesian Continual Learning} Further for incorporating the new task into the posterior, the posterior $p(\theta |\textbf X_t^{train})$ is used as the prior of the next task:
\begin{equation}
\begin{split}
\label{eq13333}
p(\theta |{\textbf X_t^{train}},{\textbf X_{t+1}^{train}}) = \frac{p({\textbf X_{t+1}^{train}}|\theta )p(\theta |{\textbf X_t^{train}})}{{p(\textbf X_{t+1}^{train})}}
\end{split}
\end{equation}

Then we can test the performance of continual learning by evaluating:
\begin{equation}
\begin{split}
\label{eq133}
p(\textbf X_t^{test},\textbf X_{t + 1}^{test}|\textbf X_t^{train},\textbf X_{t + 1}^{train}) =\\ \int {p(\textbf X_t^{test},\textbf X_{t + 1}^{test}|\theta )p(\theta |\textbf X_t^{train},\textbf X_{t + 1}^{train})} d\theta
\end{split}
\end{equation}

Similarly, $p(\theta |\textbf X_t^{train},\textbf X_{t + 1}^{train})$ can be approximated by a Gaussian using Laplace approximation whose mean is the mode of the posterior:
\begin{equation}
\begin{split}
\label{eq113}
\theta _{t,t + 1}^* = \arg {\max _\theta }\log p(\theta |\textbf X_t^{train},\textbf X_{t + 1}^{train}) \\= \arg {\max _\theta }\log p(\textbf X_{t + 1}^{train}|\theta ) + \log p(\textbf X_t^{train}|\theta ) \\- \underbrace {\log p(\textbf X_{t + 1}^{train})}_{const.}
\end{split}
\end{equation}

\section{Method}
\begin{figure*}[htb] 
\centering 
\includegraphics[height=8cm,width=16cm]{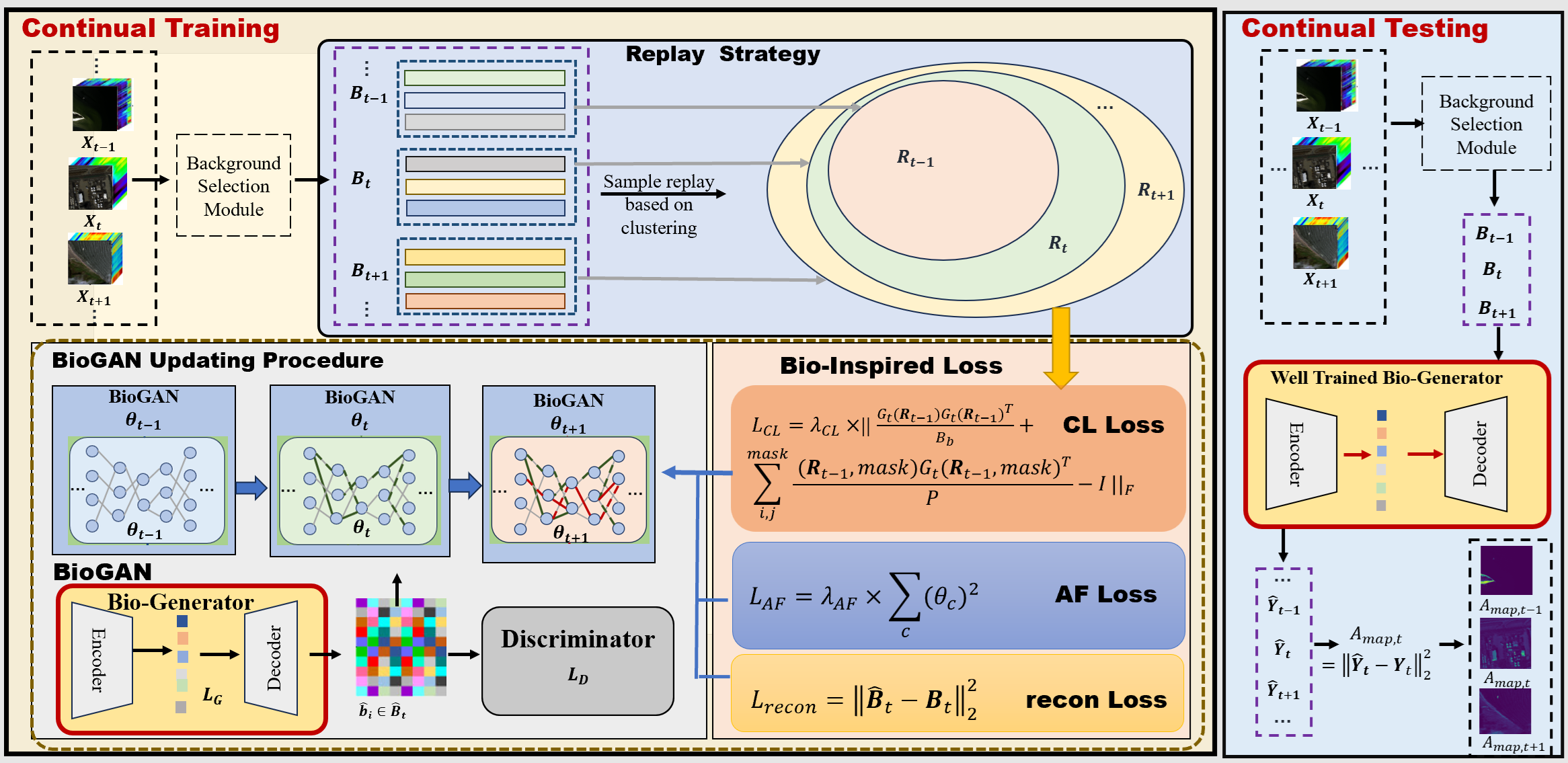} 
\caption{Overview of the proposed CL-BioGAN for OHAD. The left part represents the continual training phase, which mainly composed with Replay Strategy, BioGAN structure and Bio-Inspired Loss (CL Loss, AF Loss and recon Loss).
The right part in the figure is continual testing part, the well-trained Bio-Generator is only used in testing phase to detect the continually arrived HAD scenarios.
}
\label{overview} 
\end{figure*}

In this section, we mainly introduce the proposed CL-BioGAN by combining knowledge learned from history and present tasks. The overall architecture of the proposed CL-BioGAN is represented in Fig. \ref{overview}, which is mainly composed of CL structure and BioGAN networks.

Suppose that there are $T$ different scenario domains $\textbf{{X}}_{1},...,\textbf{X}_t,...,\textbf{X}_T$ for OHAD task datasets, each $\textbf{{X}}_{t}$ represents the $t$-th HAD scenario task. A HSI can be represented as $\textbf{X}_t\in\mathbb{R}^{M\times N\times C}=\{\textbf{x}_{i}\in\mathbb{R}^{C}\}_{i=1}^{i=M\times{N}}, t\in{1,2,…,T}$, where $\textbf{x}_{i}$ represents the $i$-th spectral vector in $\textbf{X}_t$.

\subsection{The structure of BioGAN}
The whole BioGAN for OHAD is mainly consists of three components: the background selection module (BSM), the Generator module $G$ and the Discriminator module $D$.

1) Background Selection Module (BSM): BSM is shown in Fig.\ref{overview} (b), which consists of the two main process: a) Spectral-Based Background Selection Strategy(SBSS), and b) Spatial-Spectral Noise Smoothing Strategy (SSNS). The pseudo-label for the background samples and anomalies are first coarsely selected through SBSS. Then, SSNS is used to involve spatial information for better exploiting and enhancing spectral-spatial features for each vector.

\paragraph{Spectral-Based Background Selection Strategy (SBSS)} Aim to approximately obtain the background datasets, we calculate the similarity value of neighboring spectral vectors based on SAM \cite{yuhas1992discrimination} for coarsely selecting the background sample sets $\textbf{B}_t$ for any HAD scenario $\textbf{{X}}_{t}$. 
\begin{equation}
\begin{split}
\label{eq3}
SAM{\left(\textbf{x}_{i},\textbf{x}_{j}\right)}=\frac{\textbf{x}_{i}\cdot \textbf{x}_{j}}{\left|\left|\textbf{x}_{i}\right|\right|\times\left|\left|\textbf{x}_{j}\right|\right|}
\end{split}
\end{equation}
\begin{equation}
\begin{split}
\textbf{s}_{i}=\left\{
\begin{aligned}
\label{eq4}
0 & , & SAM{\left(\textbf{x}_{i},\textbf{x}_{j}\right)}\geq \mu, \\
1 & , & SAM{\left(\textbf{x}_{i},\textbf{x}_{j}\right)}< \mu.
\end{aligned}
\right.
\end{split}
\end{equation}
\begin{equation}
\begin{split}
\label{eq8}
\textbf{B}_t =\{\textbf{index}_{(i)}\{\textbf{s}_{i}==0\}\}
\end{split}
\end{equation}

The threshold $\mu$ is defined to determine whether $\textbf{x}_{i}$ is a background sample or an anomalous sample. The value of $\textbf{s}_{i}=0$ indicates that the current pixel is part of the background, whereas $\textbf{s}_{i}=1$ indicates the current pixel is abnormal pixel, where $\textbf{B}_t \in\mathbb{R}^{n_b\times C }$ and ${n}_b$ represents the number of background.

\paragraph{Spatial-Spectral Noise Smoothing Strategy (SSNS)} We developed SSNS to incorporate spatial pixel information into spectral data, which concatenate the original spectral vector values with the average value of the spectral vectors of the $w\times{w}$ local region around the center pixel (in the experiment, we take 
$w$ as 3), and compute the mean vector as.
\begin{equation}
\begin{split}
\label{eq5}
\bar{\textbf{x}}_{i}=\frac{1}{w\times w}\sum_{r=1}^{w^2}\textbf{x}_{i}^r
\end{split}
\end{equation}
where $r$ is the pixel's index in the $w\times{w}$ region. The original spectral vector $\textbf{x}_{i}$ and the mean vector $\bar{\textbf{x}}_{i}$ are merged to create the enhanced vector $\textbf{e}_{i}$:
\begin{equation}
\begin{split}
\label{eq6}
\textbf{e}_{i}=\textbf{x}_{i}\otimes\bar{\textbf{x}}_{i}
\end{split}
\end{equation}
where $\textbf{e}_{i}\in\mathbb{R}^{2C}$ and $\otimes$ represents the concatenating operation across the channel dimension. Therefore, the augmented spectral-spacial background sample sets $\textbf{B}_t$ updated as.
\begin{equation}
\begin{split}
\label{eq88}
\textbf{B}_t \in\mathbb{R}^{n_b\times 2C }=\{\textbf{e}_{i}\vert \ \textbf{index}_{(i)}\{\textbf{s}_{i}==0\}\}
\end{split}
\end{equation}

2) Generator Module $G$: Further for improving reconstruction performance of background pixels, we introduce an advanced Encoder-Decoder architecture, where the structure of $G$ is shown in Fig. \ref{overview} (b). The AE \cite{80} has been widely used in various applications such as HSI unmixing \cite{81} and target detection \cite{82} because of the simple and outstanding performance in feature extraction. For the simplest AE structure, in which the encoder and decoder are both fully connected layer. Therefore, for the arbitrary selected background input $\textbf{b}_i \subset \textbf{B}_t \in\mathbb{R}^{n_b\times 2C}$ and the hidden layer $\textbf{z}_{i} \subset \textbf{Z}_t \in\mathbb{R}^{n_b\times L}$ with $L (L < 2C)$ dimensions, the reconstructed output $\hat {\textbf {b}}_i \subset \hat {\textbf{B}}_t \in\mathbb{R}^{n_b\times 2C}$ can be calculated as follows:
\begin{equation}
\begin{split}
\label{eq9}
\textbf{z}_{i} = f({\textbf{W}_{en}}{\textbf{b}_i} + \textbf{bias}_{en})
\end{split}
\end{equation}
\begin{equation}
\begin{split}
\label{eq10}
 \hat {\textbf {b}}_i = f({\textbf{W}_{de}}{\textbf{z}_i} + \textbf{bias}_{de})
\end{split}
\end{equation}
where $\textbf{W}_{en}$ and $\textbf{bias}_{en}$ and $\textbf{W}_{de}$ and $\textbf{bias}_{de}$ denote the weights and biases of the encoder and decoder, respectively, and $f(\cdot)$ is an inconclusive activation function. Based on the reality
that the majority of pixels in HSIs belong to the background while only a few pixels are anomalies, AE mainly progressively emphasizes the background features during the training process, and finally reconstructed background pixels as the inputs background pixels as similar as possible. Therefore, the training goal of the proposed BioGAN is to learn the features of training dataset ${\textbf {b}}_i$ and reconstruct $\hat {\textbf {b}}_i$ as similar as possible. The reconstruction error of a well-trained AE can be used to calculate the final anomaly detection result.

3) Discriminator Module $D$:
$D$ is shown in Fig. \ref{overview} (b), which is used to learn more complicated spectral properties as well as support the training of $G$. $D$ consists of two modules: the multi-scale convolution (MSC) module for extracting multi-scale features, and the $\textbf L_{2}$ SA ($\textbf L_{2}$ norm Self-Attention) module for channel feature extraction and channel correlation. Three multi-scale spectral feature maps $\textbf{F}$ have been generated by using $1\times 1$, $3\times3$, and $5\times 5$ convolution operation. Meanwhile, residual connection structure is utilized in $\textbf L_{2}$ SA module, and feature information in different levels can be extracted by concatenating the feature maps of several receptive fields together, and Spatial Position Embedding is introduced to capture the local discrimination details and the general characteristic changes of the spectral information. For an arbitrary input $\hat {\textbf {B}}_t \in\mathbb{R}^{n_b\times 2C}$, the corresponding output of an SA block can be formulated as follows.
\begin{equation}
\begin{split}
\label{eq55}
\textbf Q = {\textbf{W}_q}{{\textbf {F}}} \in {\mathbb{R}^{{n_b} \times L}}\\
\textbf K = {\textbf{W}_k}{{\textbf {F}}} \in {\mathbb{R}^{{n_b} \times L}}\\
\textbf V = {\textbf{W}_v}{{\textbf {F}}} \in {\mathbb{R}^{{n_b} \times L}}
\end{split}
\end{equation}
\begin{equation}
\begin{split}
\label{eq555}
\textbf{SA} = {Attention}(\textbf{Q},\textbf{K},\textbf{V}) = softmax(\frac{{||\textbf{Q}|{|_2}||\textbf{K}|{|_2}^T}}{{\sqrt {{2C}} }}) \textbf{V} 
\end{split}
\end{equation}
where $\textbf{W}_q \in {\mathbb{R}^{{2C} \times L}}$, $\textbf{W}_k \in {\mathbb{R}^{{2C} \times L}}$ and $\textbf{W}_v \in {\mathbb{R}^{{2C} \times L}}$ are three predefined transformation matrices, respectively. $|| \cdot |{|_2}$ represents the $\textbf L_{2}$ norm of $\textbf{SA} \in {\mathbb{R}^{{n_b} \times L}}$. We can observe from equation (\ref{eq55}) and (\ref{eq555}) that the $\textbf{SA}$ considers the relationships between each input and all other inputs. Especially for HSIs, the $\textbf{SA}$ can fully excavate more important spatial information for Background reconstruction.

A linear full-connection layer is utilized for converting the intermediate feature dimension to the output feature dimension $2C$, after the completion of the $\textbf L_{2}$ SA computation. The following final attention feature map is caculated as follow:
\begin{equation}
\begin{split}
\label{eq13}
\hat {\textbf{F}}= FFN((Linear(\textbf{SA}) + \textbf{F})) + (Linear(\textbf{SA})+ \textbf{F})
\end{split}
\end{equation}
where $\hat {\textbf{F}}\in\mathbb{R}^{{n_b}\times{2C}}$ and $FFN$ represents Feedforward Neural Network, which is consisted with Dropout, Linear, and the stable GELU activation function. Then, $\hat {\textbf{F}}$ is imported to the a multiple layer perception (MLP) to obtain the final discrimination result.

\subsection{Sample replay based on clustering}
After $t$ task has been learned, in order to alleviate catastrophic forgetting problem, an alternative sampling approach for sample replay selection is designed by involving sample replay strategy to random choose and store pixels from historical data. Suppose the background set $\textbf B_t$ with totally ${M}_t$ samples, we cluster the whole data set into $P$ groups by utilizing K-Means technique, Overall, the Expectation-Maximization (EM) optimization algorithm is used \cite{EM}, and the optimization objective function is as follows:
\begin{equation}
\begin{split}
\label{eqKM}
E =\sum_{i=1}^P{\sum_{x\in C_i}{\left\| x-\mu_i \right\| ^2}}, 
\end{split}
\end{equation}

where $C_i$ represents the set of all sample points of each cluster, $x$ denotes a sample point belonging to $C_i$, and $\mu_i$ represents the center of the 
$i$\_th cluster. Each group will contain ${M}_{t}^{i} (i=1,2,...,P)$ samples. We assume that each task adaptively selects $N$ representative samples by taking into account memory restriction. We take the sum of one percent of ${M}_{t}^{i}$ closest to the sample center in each cluster, $N=\sum_{i=1}^P{\left( M_{t}^{i} \times 100 \right) /10000}$ denoted as $N$. The specific value of $N$ is detailed in the experimental section.
the sample replay equation can be represented as
\begin{equation}
\begin{split}
\label{eq1}
k\left(\textbf B_t \right)=\bigcup_{i=1}^{P}{K_{Means}\left[:\left\lfloor\frac{{N\ast {M}_{t}^{i}}}{M_t}\right\rfloor\right]}, 
\end{split}
\end{equation}
where samples in $K_{Means}$ have been organized in ascending order based on their distance from each cluster's center pixel, and $\frac{{N\ast {M}_{t}^{i}}}{M_t}$ is the picked data from each cluster sets. The computed representative sample set of the $i$-th cluster is represented by the symbol $K_{Means}\left[:\left\lfloor\frac{{N\ast {M}_{t}^{i}}}{M_t}\right\rfloor\right]$. The replay buffer may be modified as follow.
\begin{equation}
\begin{split}
\label{eq2}
\textbf{R}_t\gets\ \textbf{R}_{t-1}\cup\ k\left(\textbf B_t\right),\textbf{R}_{t-1}=\phi\ when\ t=1
\end{split}
\end{equation}

We may lessen the effects of the imbalanced data distribution and increase the spectral anomaly detection algorithm's resilience by using equation (\ref{eq2}) for adaptive replay technique. Obviously, abnormalities in various HSI sets may be continually detected with better accuracy performance based on the clustering sample replay in CL module.

\subsection{CL Bio-inspired loss function}
In this part, we mainly introduce the proposed Bio-Inspired loss of CL based on the Bayesian learning framework, which hypothesized based on biological synaptic plasticity by tracking the probability distribution of synaptic weights under dynamic sensory inputs \cite{7}. 
The whole proposed CL Bio-Inspired Loss function is composed with 1) ACL loss $L_{ACL}$ and 2) the BioGAN loss $L_{BioGAN}$. The ACL loss is used to assure stability for detection and prevent catastrophic forgetting for cross-domain scenarios. BioGAN loss mainly ensures the training fidelity performance for background sample reconstruction. Therefore, we construct and summarize the whole loss function as follow:
\begin{equation}
\begin{split}
\label{eq200}
L = L_{BioGAN} + L_{ACL}
\end{split}
\end{equation}
\begin{equation}
\begin{split}
\label{eq222}
L_{BioGAN} = {L_G} + {L_D} + {L_{recon}}
\end{split}
\end{equation}
\begin{equation}
\begin{split}
\label{eq2222}
L_{ACL} = {L_{CL}} + {L_{AF}}
\end{split}
\end{equation}

1) BioGAN loss:
The spatial-spectral background sets $\textbf{B}_t$ and reconstruction background set $\hat{\textbf{B}}_t = G(\textbf{B}_t)$ is measured by using the mean square error (MSE) loss:

\begin{equation}
\begin{split}
\label{eq17}
L_{recon} =  \left\| {\left. {{\hat{\textbf{B}}_t} - {\textbf{B}_t}} \right\|} \right._2^2
\end{split}
\end{equation}

We performed differentiable data enhancement ($Diff$) on the data and pseudo-data in the $G$ and $D$ to assure the training stability of GAN without appearing feature shift phenomenon for brightness, saturation, and contrast information to realize better convergence results. Consequently, the adversarial process allows the $G$ and $D$ to be updated alternately as:
\begin{equation}
\begin{split}
\label{LG}
{L_G} = {{\rm E}_{\textbf{x} \sim p(\textbf{B}_t)}}[\log (1 - D(Diff(G(\textbf{x}))))]
\end{split}
\end{equation}
\begin{equation}
\begin{split}
\label{eq19}
{L_D} = -{E_{\textbf{x} \sim p(\textbf{B}_t \cup \hat {\textbf{B}}_t)}}[\log (D(\textbf{x})) +  \log (1 - D(Diff(G(\textbf{x}))))]
\end{split}
\end{equation}


2) ACL loss: ACL loss is composed of CL loss to prevent weight drift caused by new knowledge and the AF loss for selectively converging the related parameter weights of the network. 
\paragraph{AF loss} Inspired by biological forgetting and Bayesian CL, we introduce a forgetting factor $\beta$ and replace $p(\theta |{\textbf X_t^{train}})$ that absorbs all the information of ${\textbf X_t^{train}}$ with a weighted product distribution.
\begin{equation}
\begin{split}
\label{eq144}
\hat p(\theta |{\textbf X_t^{train}},\beta ) = \frac{{p{{(\theta |{\textbf X_t^{train}})}^{(1 - \beta )}}p{{(\theta )}^\beta }}}{Z}
\end{split}
\end{equation}
where we denote that $p(\theta |{\textbf X_t^{train}})$ and $p(\theta)$ are mixed to produce the new distribution $\hat p$. $Z$ is the normalization value that depends on $\beta$, which keeps $\hat p(\theta |{\textbf X_t^{train}},\beta )$ following a Gaussian distribution if $p(\theta |{\textbf X_t^{train}})$ and $p(\theta)$ are both Gaussian. When $p(\theta)$→0, $\hat p$ will be dominated by $p(\theta |{\textbf X_t^{train}})$ and remember all the information about task $t$, whereas $p(\theta)$→1, $\hat p$ will actively forget all the information about task $t$. By modifying Eq.(\ref{eq133}), the new target becomes
\begin{equation}
\begin{split}
\label{eq50}
p(\textbf X_t^{test},\textbf X_{t + 1}^{test}|\textbf X_t^{train},\textbf X_{t + 1}^{train}, \beta) =\\ \int {p(\textbf X_t^{test},\textbf X_{t + 1}^{test}|\theta )p(\theta |\textbf X_t^{train},\textbf X_{t + 1}^{train}, \beta)} d\theta
\end{split}
\end{equation}

where $\beta$ decides how much information from task $t$ should be forgotten to maximize the probability of learning task $t+1$. A suitable $\beta$ value should be defined as follows
\begin{equation}
\begin{split}
\label{eq51}
{\beta ^*} = \arg {\max _\beta }p(\textbf X_{t + 1}^{train}|\textbf X_t^{train},\beta ) \\= \arg {\max _\beta }\int {p(\textbf X_{t + 1}^{train}|\theta )\hat p(\theta |\textbf X_t^{train},\beta )d\theta }
\end{split}
\end{equation}
Here, $\beta$ should be between 0 and 1, we make a grid search to determine the optimal value for $\beta$. Next, $p(\theta |{\textbf X_t^{train}}, {\textbf X_{t + 1}^{train}},\beta)$ can also be estimated by the approximated Gaussian using Laplace approximation, and the Maximum-a-Posteriori Estimation (MAP) is:
\begin{equation}
\begin{split}
\label{eq52}
\theta _{t,t + 1}^* = \arg {\max _\theta }\log p(\theta |\textbf X_t^{train},\textbf X_{t + 1}^{train},\beta ) \\= \arg {\max _\theta }(1 - \beta )(\log p(\textbf X_{t + 1}^{train}|\theta ) \\+ \log p(\theta |\textbf X_t^{train})) + \beta \log p(\theta |\textbf X_{t + 1}^{train}) + const.
\end{split}
\end{equation}
Then, we obtain the active forgetting AF loss function by synaptic Expansion-Convergence (AF loss).
\begin{equation}
\begin{split}
\label{eq16}
{L_{AF}} = {\lambda _{AF}} \times \sum\limits_c {{{({\theta _c})}^2}} 
\end{split}
\end{equation}
where ${{\theta _c}}$ represents the parameters of the current task. Fig. \ref{tuchu} (b) indicates the implementation of active forgetting in a CL model.

\paragraph{CL loss} This loss function minimizes the magnitude of parameter updates by forcing the distance between the input data of task $t$ and the reconstructed image of task $t+1$ to get closer to preserve stability for OHAD. 
We use a replay sample set $\textbf{R}_{t}$ to reduce the distance between different task models for stability protection in continual learning. 
Lastly, the definition of the F-norm regularization loss for every task $t$ is:
\begin{equation}
\begin{split}
\label{eq14}
{L_{CL}} = \lambda_{CL}  \times \left\| {\frac{{{G_t}(\textbf{R}_{t - 1}){G_t}{{(\textbf{R}_{t - 1})}^T}}}{B_b}} \right. + \\
{\left. {\sum\limits_{i,j}^{mask} {\frac{{{G_t}(\textbf{R}_{t - 1,mask}){{{G_t}(\textbf{R}_{t - 1,mask})}^T}}}{P}}  - \textbf{I}} \right\|_F}
\end{split}
\end{equation}

Among them, $B_b$ stands for batch-size, $P$ denotes the number of pixel in $mask$. $\textbf{I}$ stands for the identity matrix, ${\left\| {\left.  \bullet  \right\|} \right._F}$ represents the matrix's Fibonacci norm, and $\lambda _{CL}$ represents the balancing parameter. The CL loss will gradually update the parameters of the Generator in accordance with the continuously coming tasks.

\subsection{Continual Anomaly Detection} During the testing procedure in Fig. \ref{overview} (c), the final projected anomaly probability map is obtained by subtracting the continuous data $\textbf{{Y}}_{1},...,\textbf{Y}_t,...,\textbf{Y}_T$ from the reconstructed background pixel $\hat {\textbf{{Y}}}_{1} = G(\textbf{{X}}_{1}),..., \hat {\textbf{{Y}}}_{t} = G(\textbf{{X}}_{t}),..., \hat {\textbf{{Y}}}_{T} = G(\textbf{{X}}_{T})$ by our proposed well-trained CL-BioGAN.
\begin{equation}
\begin{split}
\label{eq20}
A_{map,t} = \left\| {\left. {\hat{\textbf{Y}}_t - \textbf{Y}_t} \right\|_2^2} \right.
\end{split}
\end{equation}

The detail pseudo-code for the proposed CL-BioGAN is introduced in Algorithm \ref{alg:algorithm}.

\begin{algorithm}[tb]
    \caption{CL-BioGAN}
    \label{alg:algorithm}
    \textbf{Input}: The stream HSIs $\textbf{X}_1$,$\ldots$,$\textbf{X}_t$,$\ldots$,$\textbf{X}_T$ for $T$ tasks.\\
    \textbf{Parameter}: $\mu=0.99$, the number of clustering groups $P=3$, $w=3$, the channel dimension $C=dmin(\textbf{X}_1,..., \textbf{X}_T)$ , $\lambda _{AF}=0.1$, $\lambda _{CL}=0.9$, $lr=0.00005$.\\
    \textbf{Output}: $A_{map,t}$.
    \begin{algorithmic}[1] 
         \STATE Initialize replay buffer $\textbf{R}_1$=$\phi$, the model parameters $\theta_1$.
         \FOR{ $t$=1 to $T$ tasks }
	{
		\STATE $\textbf{X}_t\in\mathbb{R}^{M\times N\times C}={\{\boldsymbol{x}_{i}\in\mathbb{R}^{C}\}}_{i=1}^{i=M\times N}$;
		
		\FOR {$H$ training epochs}
		{
            \STATE generate $\textbf{s}_{i}$ vector by (\ref{eq3}) and (\ref{eq4});
            
           \STATE construct background samples set $\textbf{B}_t$ by (\ref{eq88});
            
            \IF  {$t$ = 1} 
            \STATE update $\theta_1$ by minimizing $L_G$, $L_D$ and $L_{recon}$ through equation (\ref{LG}), (\ref{eq19}) and (\ref{eq17}), as equation (\ref{eq16}) $L_{AF}=0$, equation (\ref{eq14}) ${L_{CL}=0}$;
            \ELSE 
            \STATE update $\theta_t$ by minimizing $L_G$, $L_D$ and $L_{recon}$ through equation (\ref{LG}), (\ref{eq19}) and (\ref{eq17});
            \ENDIF
            
            \STATE apply K-Means to obtain $M_i\left(i=1,2,3\right)$ containing three groups exemplars respectively for the subset through (\ref{eq1});
            
            \IF  {$t$ = 1} 
            \STATE update replay buffer by $\textbf{R}_1\ \gets\ k\left(\textbf{B}_t \right)$ ;
            
            \ELSE 
            \STATE update replay buffer by $\textbf{R}_t\ \gets \textbf{R}_{t-1}\cup\ k\left(\textbf{B}_t \right)$;
            \ENDIF
		}
       \ENDFOR
       
	}
       \ENDFOR
       \\
       \STATE Construct the detection for tasks through equation (\ref{eq20});
    \end{algorithmic}
\end{algorithm}

\section{Experiences}
In this section, we mainly assess the proposed CL-BioGAN for open scenario OHAD on five cross-domain distinct hyperspectral data sets. The following part we will introduce and analyze the performance of proposed CL-BioGAN from experimental setting, detection results and ablation experiment under open scenario circumstance.

\subsection{Experiment Setting}

\textbf{1. Datasets and Computing Infrastructure:} We evaluated our method on two types of datasets: ABU-Urban and ABU-Beach (ABU) \cite{52}, as well as HAD100 for OHAD \cite{91}. For the ABU dataset, we select five different real HSIs to form the CL tasks: Los Angeles-1, Bay Champagne, Los Angeles-2, Cat Island and San Diego. These HSIs are captured by the Airborne Visible/Infrared Imaging Spectrometer (AVIRIS) sensor under different scenarios with distinct spatial resolutions, channels, spatial size, spectral bands. The detailed information of these datasets is illustrated in Table \ref{Data}. Meanwhile, we random select five different areas from HAD100 dataset as shown in Table \ref{HAD10Data} as ang20170821t183707\_91, ang20170821t183707\_100, ang20191004t185054\_24, ang20210614t141018\_33, ang20170908t225309\_40, are respectively denoted as HAD100\_91, HAD100\_100, HAD100\_24, HAD100\_33, HAD100\_40, which are all captured by Airborne Visible/Infrared Imaging Spectrometer-Next Generation (AVIRIS-NG).
All the experiments conducted under the hardware environment: Intel(R) Xeon(R) CPU at 2.30 GHz with 64-GB RAM and an NVIDIA GeForce GTX 2080 Ti graphical processing unit (GPU) with 11-GB RAM, and a 64-bit Windows 10 system and the Pytorch 1.10.2 DL frameworks. 

\begin{table}[!ht]
	\centering
	\caption{Details Of The ABU Anomaly Detection Data Set}
	\label{Data}
	\setlength{\tabcolsep}{0.5mm}{
	\begin{tabular}{cccccc}
		\toprule  
		HSIs & Spatial size &	channels	& resolution/m & bands/nm &	Sensor \\
		\midrule  
		Los Angeles-1 &	100$\times$100 &	205 &	7.1	& 430 - 860 &	AVIRIS \\
  \midrule  
		Bay Champagne &	100$\times$100 &	188	& 4.4 &	400 - 2500 &	AVIRIS \\
		\midrule  
		Los Angeles-2 &	100$\times$100 &	205 &	7.1 &	430 - 860 &	AVIRIS \\
		\midrule  
		Cat Island &	150$\times$150 &	188 &	17.2 &	400 - 2500 &	AVIRIS \\
		\midrule  
		San Diego &	100$\times$100 &	193 &	7.5 &	400 - 2500 &	AVIRIS \\
		\bottomrule  
	\end{tabular}
	}
	
\end{table}

\begin{table}[!ht]
	\centering
	\caption{Details Of The HAD100 Anomaly Detection Data Set}
	\label{HAD10Data}
	\setlength{\tabcolsep}{0.5mm}{
	\begin{tabular}{cccccc}
		\toprule  
		HSIs & Spatial size &	channels	& resolution/m & Date &	Sensor \\
             \midrule  
		HAD100\_91 &	64$\times$64 &	276 &	2.3 &2017/8/21	 &	AVIRIS-NG \\
         \midrule  
		HAD100\_100 &	64$\times$64 &	276 &	2.3 &	2017/8/21 &	AVIRIS-NG \\
         \midrule  
		HAD100\_24 &	64$\times$64 &	276 &	8.4 &	2019/10/4&	AVIRIS-NG \\
        \midrule  
		HAD100\_33 &	64$\times$64 &	276 &	1.9 &	2021/6/14&	AVIRIS-NG \\
         \midrule  
		HAD100\_40 &	64$\times$64 &	276 &	2.0 &	2017/9/8&	AVIRIS-NG \\
		
		\bottomrule  
	\end{tabular}
	}
	
\end{table}

\textbf{2. Methods:}
In order to demonstrate the superiority of our proposed CL-BioGAN for OHAD, we select the most representative methods from each paradigm to establish the benchmark, these training methods include DL based Joint(J) training for OHAD, Fine-tune(FT) based OHAD, and CL-based algorithm MAS\cite{53}, EWC\cite{54}, OWM\cite{55}, and CL-CaGAN \cite{60}.

\textbf{3. Metrics:} In this paper, we mainly utilize receiver operating characteristics (ROC) \cite{zweig1993receiver} and the area under the ROC curve (AUC) \cite{ferri2011coherent} to quantitatively assess the performance of detectors. 
The ROC curve illustrates the relationship between true positive rate ($P_D$) and false positive rate ($P_F$). 
A 3D ROC curve can be generated by a triplet parameter vector specified by ($P_D$, $P_F$, $\tau$), or by three 2D ROC curve of ($P_D$, $P_F$,), ($P_D$, $\tau$)
and ($P_F$, $\tau$) and their AUC values expressed as ${AUC}_{(D,F)}$, ${AUC}_{(D,\tau)}$ and ${AUC}_{(F,\tau)}$. In the experiments, we use ${AUC}_{(F,\tau)}$ and ${AUC}_{BS}$ \cite{chen2021component} to represent the background suppression(BS) rate, and ${AUC}_{(D,F)}$
are used to evaluate the performance of the detector. The ${AUC}_{BS}$ can be computed by ${AUC}_{(D,F)}$ and ${AUC}_{(F,\tau)}$:
\begin{equation}
\begin{split}
\label{eq18}
{AUC}_{BS} = {{AUC}_{(D,F)} - 
 {AUC}_{(F,\tau)}}
\end{split}
\end{equation}

The accuracy ($ACC$) as the average test ${AUC}_{(D,F)}$ of all tasks is utilized to better assess the capacity of CL to identify anomalies. Meanwhile, Backward transfer ($BWT$) represents the average forgetting measure of the model after completing $T$ tasks.
In order to explore the learning plasticity of new tasks in OHAD, we also use the Forward Transfer ($FWT$) to continuously estimate the learning ability for the new task. 
The following are the fomulas of $ACC$, $BWT$ and $FWT$:
\begin{equation}
\begin{split}
\label{eq21}
ACC = \frac{1}{T}\sum\limits_{i = 1}^T {AU{C_{(D,F)T,i}}}
\end{split}
\end{equation}
\begin{equation}
\begin{split}
\label{eq221}
BWT = \frac{1}{{T - 1}}\sum\limits_{i = 1}^{T - 1} {(AU{C_{(D,F)T,i}} - AU{C_{(D,F)T - 1,i}})} 
\end{split}
\end{equation}
\begin{equation}
\begin{split}
\label{eq22}
FWT = \frac{1}{{T - 1}}\sum\limits_{i = 2}^T {(AU{C_{(D,F)T,i}} - AU{C^*}_{(D,F)i})} 
\end{split}
\end{equation}

The $AU{C^*}_{(D,F)i}$ represents the result of training on the $i$-th task. The $AUC_{(D,F)T,i}$ represents the $i$-th task after learning the $T$-th task, where $T$ represents the total number of the tasks, and $i$ refers to the $i$-th task being evaluated. During the inference procedure, we evaluate the model after training on all tasks. 

\textbf{4. The number of $N$:} For the sample replay strategy, the number of samples in each cluster is $P$=1, $P$=2, $P$=3, and the specific value of the total replay samples $N$ is selected adaptively, which is shown in Table \ref{N}.

\begin{table}[!ht]
	\centering
     
	\caption{The number of samples in each cluster (total number and selected number) and the specific value of $N$ in the sample replay.}
	\label{N}
	\setlength{\tabcolsep}{0.5mm}{
	\begin{tabular}{cccccccc}
		\toprule  
	\multirow{2}{*}{HSIs} &  \multicolumn{2}{c}{$P$=1}&\multicolumn{2}{c}{$P$=2} & \multicolumn{2}{c}{$P$=3} & \multirow{2}{*}{$N$}\\
    & Total & Selected & Total & Selected & Total & Selected& \\
        \midrule  
Los Angeles-1    & 286 &2   & 4603&46 & 5111&51  & 99  \\
\midrule  
Bay Champagne          & 6603&66 & 2272&22 & 1125&11  & 99  \\
\midrule  
Los Angeles-2          & 4919 &49& 2263&22 & 2818&28  & 99  \\
\midrule  
Cat Island             & 364 &3  & 3531&35 & 18605&186 & 224 \\
\midrule  
San Diego              & 529 &5  & 8858&88 & 613&6   & 99  \\
\midrule  
HAD100\_91             & 1734&17 & 886&8& 1476&14  & 39  \\
\midrule  
HAD100\_100            & 1243&12 & 676&6  & 2177&21  & 39  \\
\midrule  
HAD100\_24 & 769&7& 952&9& 2375&23 & 39  \\
\midrule  
HAD100\_33             & 161&1  & 2432&24 & 1503&15  & 40  \\
\midrule  
HAD100\_40             & 2348&23 & 158&1  & 1590&15  & 39  \\
		\bottomrule  
	\end{tabular}
	}
\end{table}

\begin{table*}[ht]
	\centering
 \caption{Continual learning for OHAD performance with three evaluation metrics.}
	\begin{tabular}{ccccccccc}
		\toprule  
	ABU Dataset&	\multirow{2}{*}{Metrics}& \multicolumn{7}{c}{Method} \\
Tasks & & J-BioGAN	& FT-BioGAN & MAS-BioGAN& EWC-BioGAN & OWM-BioGAN& CL-CaGAN& CL-BioGAN(ours) \\
		
		\midrule  
\multirow{3}{*}{2}&ACC$\uparrow$ & 0.9789 &0.9797 &0.9682 & 0.9246 &0.983 &\textbf{0.9835} &0.9763 \\
&BWT$\uparrow$                     &-      &-0.0030 &-0.0263& -0.0664 &-0.0002&\textbf{0.0008}&-0.0133\\
&FWT$\uparrow$              &-     &-0.0067 &-0.0062 &-0.0534&\textbf{-0.0028}&-0.0070        &-0.0032 \\
		\midrule  
\multirow{3}{*}{3}&ACC$\uparrow$ & 0.9456 &0.8711 &0.9627 &0.9462 &0.7862 &0.6005 &\textbf{0.9734}\\
&BWT$\uparrow$                     &-      &-0.1471 &-0.0062&\textbf{0.0125}&-0.2863&-0.5575&0.0003\\
&FWT$\uparrow$                     &-      &-0.1565 &-0.0246&-0.0290&-0.1960&-0.0064&\textbf{-0.0026}\\
		 \midrule  
\multirow{3}{*}{4}&ACC$\uparrow$ & 0.9631 &0.8783 &0.9753 &0.9563 &0.6660 &0.5559&\textbf{0.9788}\\
&BWT$\uparrow$                     &-      &-0.0309 &\textbf{0.0096}&-0.0005&-0.2267&-0.1906&0.0025\\
&FWT$\uparrow$                     &-      &-0.1029 &-0.0076&-0.0192&-0.2442&-0.0061&\textbf{-0.0022}\\
             \midrule  
\multirow{3}{*}{5}&ACC$\uparrow$ & 0.9497 &0.6625 &0.9491 &0.9388 &0.9313 &0.6337&\textbf{0.9602}\\
&BWT$\uparrow$                     &-      &-0.2655 &-0.0117&0.0002 &\textbf{0.3026} &0.0178&-0.0033\\
&FWT$\uparrow$                     &-      &-0.3754 &-0.0139&-0.0194&-0.0286&-0.0067&\textbf{0.0010}\\ 
            \midrule  
-&Training Parameters$\downarrow$&-&-&-&-&-& 1140426&\textbf{752208}\\
-&Detection Parameters$\downarrow$&-&-&-&-&-&921578&\textbf{628920}\\

		\bottomrule  

\end{tabular}

	\label{ACCBWT}
\end{table*}

\begin{figure*}[!ht] 
\centering 

\subfigure[]{
\begin{minipage}[t]{0.23\linewidth}
\centering
\hspace{-6mm}
\includegraphics[width=45mm,height=36mm]{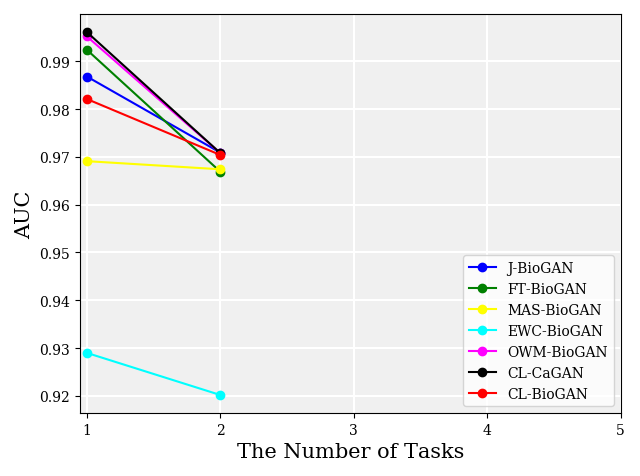}
\end{minipage}
} 
\subfigure[]{
\begin{minipage}[t]{0.23\linewidth}
\centering
\hspace{-6mm}
\includegraphics[width=45mm,height=36mm]{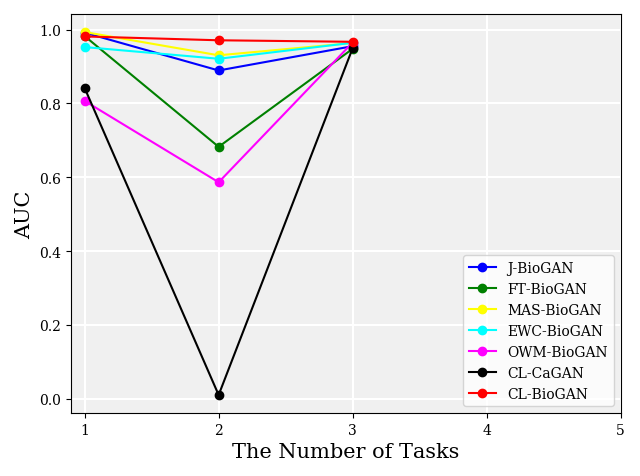}
\end{minipage}
}
\subfigure[]{
\begin{minipage}[t]{0.23\linewidth}
\centering 
\hspace{-6mm}
\includegraphics[width=45mm,height=36mm]{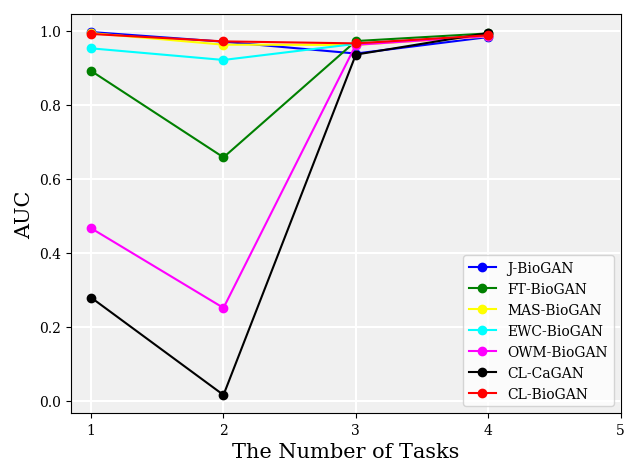}
\end{minipage}
}
\subfigure[]{
\begin{minipage}[t]{0.23\linewidth}
\centering
\hspace{-6mm}
\includegraphics[width=45mm,height=36mm]{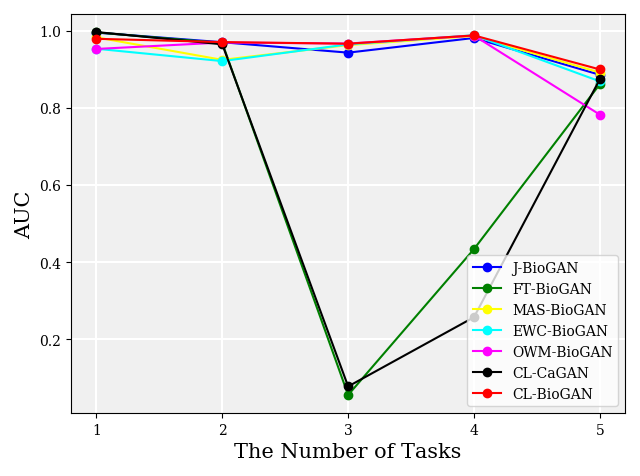}
\end{minipage}
}
\caption{The catastrophic forgetting performance evaluation for open scenario OHAD on previous tasks. (a) AUC value for tasks 1-2 after training on the second task. (b) AUC value for tasks 1-3 after training on the third task.
(c) AUC value for tasks 1-4 after training on the fourth task.
(d) AUC value for tasks 1-5 after training on the fifth task.} 
\label{AUC-change} 
\end{figure*}

\begin{figure*}[!ht] 
\centering 
\subfigure[]{
\begin{minipage}[t]{0.3\linewidth}
\centering
\hspace{-6mm}
\includegraphics[width=50mm,height=40mm]{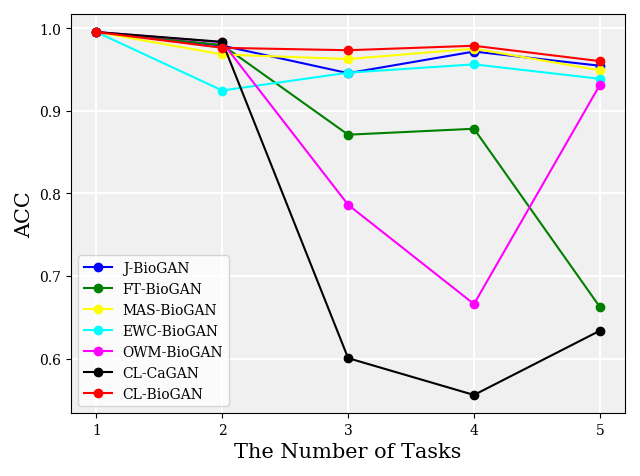}
\end{minipage}
}
\subfigure[]{
\begin{minipage}[t]{0.3\linewidth}
\centering
\hspace{-6mm}
\includegraphics[width=50mm,height=40mm]{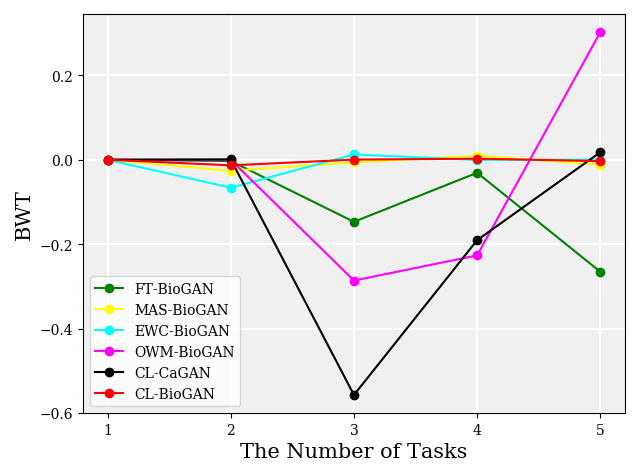}
\end{minipage}
}
\subfigure[]{
\begin{minipage}[t]{0.3\linewidth}
\centering
\hspace{-6mm}
\includegraphics[width=50mm,height=40mm]{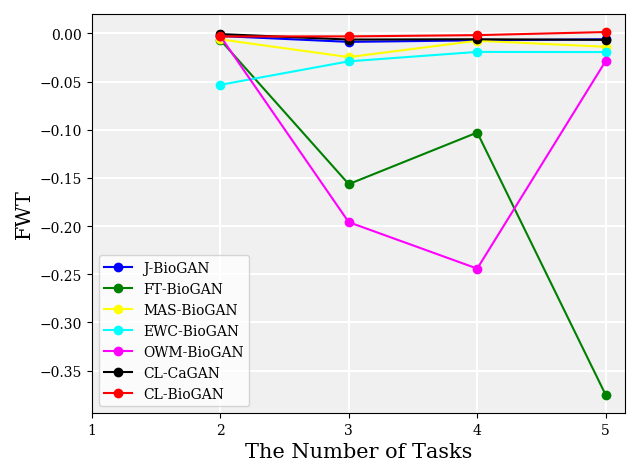}
\end{minipage}
}

\caption{The ACC, BWT and FWT Performance evaluation for OHAD of different methods. (a) The variation of $ACC$ for Task 1-5. (b) The variation
of $BWT$ for OHAD Task 1-5. (c) The variation of $FWT$ for OHAD Task 1-5.} 
\label{ACC and BWT} 
\end{figure*}

\begin{table}[ht]
	\centering
	\caption{The ${AUC}_{BS}$ impact analyses of different CL strategies. $BS$ represent the value of ${AUC}_{BS}$, the larger value means the better performance. $Avg$ represents the mean value of $BS$. The optimal results are highlighted in bold.}
	\setlength{\tabcolsep}{1.4mm}{
	\begin{tabular}{ccccccc}
		\toprule  
	ABU Dataset Tasks &1& 2 & 3  & 4 & 5 & \multirow{2}{*}{Avg$\uparrow$}\\
           Methods& ${BS}$$\uparrow$ & ${BS}$$\uparrow$& ${BS}$$\uparrow$& ${BS}$$\uparrow$& ${BS}$$\uparrow$ \\
        \midrule  
J-BioGAN&0.9938 & - & - & -& - &0.9938 \\
FT-BioGAN &\textbf{0.9952} & - & - & -& - &\textbf{0.9952} \\
MAS-BioGAN &\textbf{0.9952} & - & - & -& - &\textbf{0.9952} \\
EWC-BioGAN &\textbf{0.9952} & - & - & -& - &\textbf{0.9952}\\
OWM-BioGAN &\textbf{0.9952} & - & - & -& - &\textbf{0.9952} \\
CL-CaGAN & \underline{0.9951} & - & - & -& - & \underline{0.9951} \\
CL-BioGAN& \textbf{0.9952}& - & - & -& - &\textbf{0.9952} \\
         \midrule  
J-BioGAN & 0.9853 & \underline{0.9571} & - & -& - & 0.9712\\
FT-BioGAN & 0.9873 & 0.9535 & - & -& - & 0.9704 \\
MAS-BioGAN& 0.9682 & 0.9517 & - & -& - & 0.9599 \\
EWC-BioGAN & 0.9260 & 0.9007 & - & -& - & 0.9134 \\
OWM-BioGAN & \underline{0.9936} & \textbf{0.9574} & - & -& - & \underline{0.9755} \\
CL-CaGAN & \textbf{0.9953} & 0.9571 & - & -& - & \textbf{0.9762} \\
CL-BioGAN & 0.9805& 0.9530 & - & -& - & 0.9668 \\
         \midrule  
J-BioGAN & 0.9790 & 0.8680 & \textbf{0.9280} & -& - & 0.9250 \\
FT-BioGAN & 0.9513 & 0.6326 & 0.9162 & -& - &0.8334\\
MAS-BioGAN& \textbf{0.9921} & \underline{0.9150} & 0.9021 & -& - & \underline{0.9364}\\
EWC-BioGAN & 0.9514 & 0.9042 & 0.9001 & -& - & 0.9186 \\
OWM-BioGAN & 0.7959 & 0.5564 & \underline{0.9225} & -& - & 0.7583 \\
CL-CaGAN & -0.0932 & -0.6166 & 0.8341 & -& - & 0.0414 \\
CL-BioGAN & \underline{0.9803}& \textbf{0.9544} & 0.9018 & -& - & \textbf{0.9455} \\
         \midrule  
J-BioGAN &  \textbf{0.9949} & \textbf{0.9576} & 0.8929 & 0.9378 & - &\textbf{0.9458} \\
FT-BioGAN & 0.0691 & -0.0260 & 0.9375 & 0.9840& - & 0.4912 \\
MAS-BioGAN & \underline{0.9907} & 0.9447 & 0.8857 & 0.9304& - & 0.9379 \\
EWC-BioGAN & 0.9505 & 0.9033& \underline{0.8986} & 0.9438& - & 0.9241\\
OWM-BioGAN & 0.4403 & 0.1755 & \textbf{0.9313} & \underline{0.9671}& - & 0.6286 \\
CL-CaGAN & -0.7180 & -0.9379 & 0.7055 & \textbf{0.9694}& - & 0.0048 \\
CL-BioGAN & 0.9889& \underline{0.9534} & 0.8943 & 0.9328 & - & \underline{0.9424}\\
         \midrule  
J-BioGAN & \textbf{0.9954} & \textbf{0.9573} &\underline{0.8957} & \textbf{0.9634}& 0.8553 &\textbf{0.9334} \\
FT-BioGAN & 0.9827 & 0.9467 & -0.5055 & -0.0639& 0.8444 & 0.4409 \\
MAS-BioGAN & 0.9809 & 0.9063 & 0.8820 &0.9373& \underline{0.8613} & 0.9136 \\
EWC-BioGAN & 0.9504 & 0.9027 & \textbf{0.8971} & 0.9446& 0.8356& 0.9061 \\
OWM-BioGAN &0.9487 & \underline{0.9545} & 0.916 & \underline{0.9553}& 0.7586 &0.9066 \\
CL-CaGAN & 0.9784 & 0.9390 & -0.4341 &-0.4201& 0.8347 & 0.3796 \\
CL-BioGAN & \underline{0.9864}& 0.9526 & 0.8942 & 0.9304& \textbf{0.8625} & \underline{0.9232}\\
		\bottomrule  
	\end{tabular}
	}	

	\label{BS}
\end{table}

\begin{table}[ht]
	\centering
 \caption{Continual learning for OHAD performance with three evaluation metrics and Average ${AUC}_{BS}$ in HAD100 dataset. $Avg\ BS$ represent the value of Average ${AUC}_{BS}$}
	\begin{tabular}{cccc}
		\toprule  
	\multirow{2}{*}{HAD100 Dataset Tasks}&	\multirow{2}{*}{Metrics}& \multicolumn{2}{c}{Method} \\
 & &  CL-CaGAN& CL-BioGAN(ours) \\
		
		\midrule  
\multirow{3}{*}{2}&ACC$\uparrow$ & 0.9740 & \textbf{0.9815}  \\
&BWT$\uparrow$   &-0.0289    & \textbf{-0.0129} \\
&FWT$\uparrow$    &-0.0220    & \textbf{0.0022} \\
&$Avg\ BS$$\uparrow$    &\textbf{0.9546}    & 0.9441 \\
		\midrule  
\multirow{3}{*}{3}&ACC$\uparrow$ &0.9213 & \textbf{0.9833} \\
&BWT$\uparrow$      &-0.0861   & \textbf{0.0004} \\
&FWT$\uparrow$      &-0.1022   & \textbf{-0.0041} \\
&$Avg\ BS$$\uparrow$    &0.8940    & \textbf{0.9420} \\
		 \midrule  
\multirow{3}{*}{4}&ACC$\uparrow$ & 0.9418 & \textbf{0.9877} \\
&BWT$\uparrow$    & \textbf{0.0014}   &0.0003 \\
&FWT$\uparrow$           &-0.0629   & \textbf{-0.0026} \\
&$Avg\ BS$$\uparrow$    &0.8821    & \textbf{0.9305} \\
             \midrule  
\multirow{3}{*}{5}&ACC$\uparrow$ & 0.9595 & \textbf{0.9885}\\
&BWT$\uparrow$          & \textbf{0.0101}  &0.0001 \\
&FWT$\uparrow$          &-0.0411  & \textbf{-0.0021} \\ 
&$Avg\ BS$$\uparrow$    &0.8695    & \textbf{0.9332} \\
		\bottomrule  

\end{tabular}

	\label{HAD100ACC}
\end{table}



\begin{figure*}[!ht] 
\centering 

\subfigure[]{
\begin{minipage}[t]{0.23\linewidth}
\centering
\hspace{-6mm}
\includegraphics[width=45mm,height=35mm]{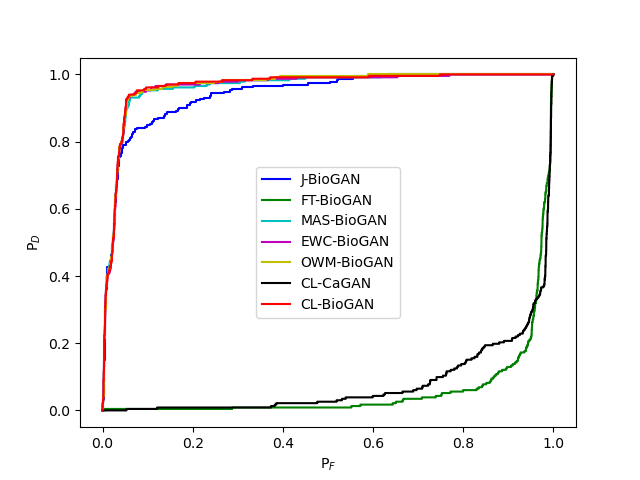}
\end{minipage}
} 
\subfigure[]{
\begin{minipage}[t]{0.23\linewidth}
\centering
\hspace{-6mm}
\includegraphics[width=45mm,height=35mm]{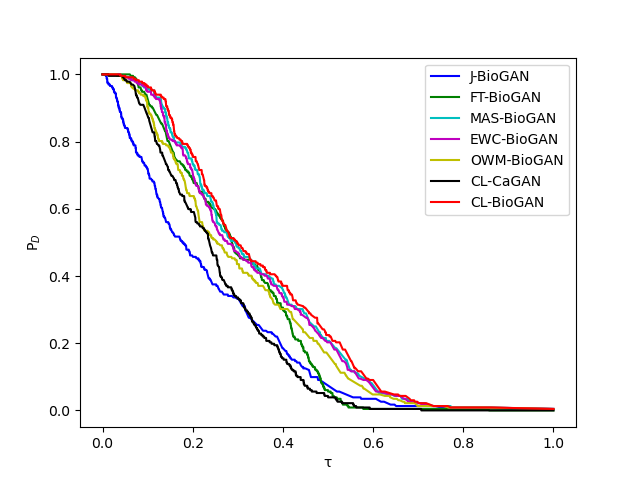}
\end{minipage}
}
\subfigure[]{
\begin{minipage}[t]{0.23\linewidth}
\centering 
\hspace{-6mm}
\includegraphics[width=45mm,height=35mm]{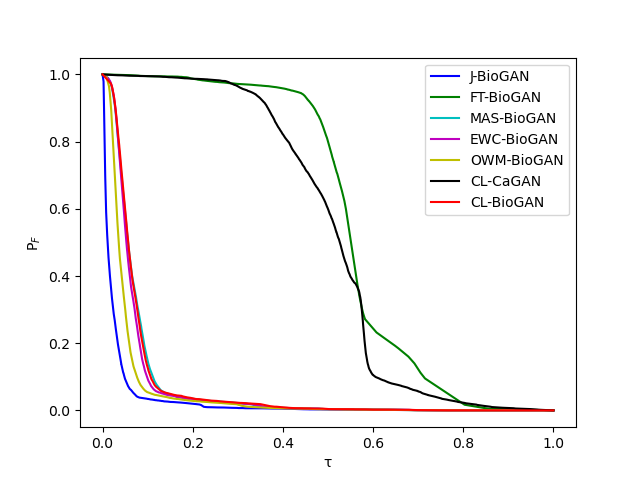}
\end{minipage}
}
\subfigure[]{
\begin{minipage}[t]{0.23\linewidth}
\centering
\hspace{-6mm}
\includegraphics[width=45mm,height=35mm]{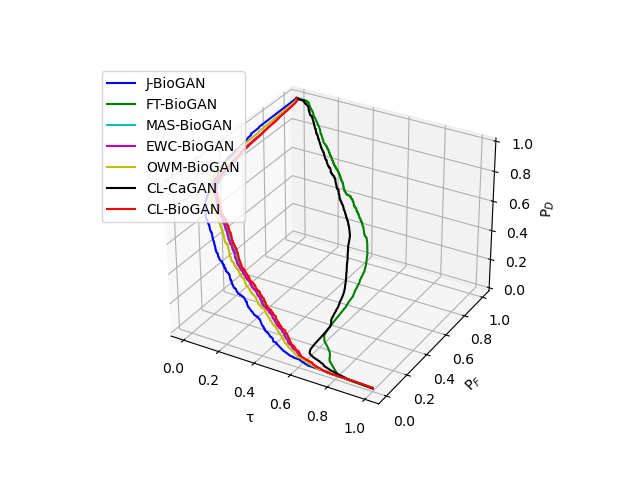}
\end{minipage}
}

\caption{The $AUC$ test results of different methods on the previous task after learning the fifth task are shown as follows: the first column (a) represents the ${AUC}_{(D,F)}$ value on Los Angeles-2 after training task 1-5. The second column (b) represents the ${AUC}_{(D,\tau)}$ value on Los Angeles-2 after training task 1-5. The third column (c) represents the ${AUC}_{(F,\tau)}$ value on Los Angeles-2 after training tasks 1-5.
(d) represents the 3D ROC value on Los Angeles-2 for tasks 1-5 after completing training on the fifth task.} 
\label{3DROC} 
\end{figure*}

\subsection{MainResults}

To verify the accuracy improvements and the robustness of our proposed CL-BioGAN for OHAD tasks under open scenario circumstance, we conducted comprehensive evaluations of the aforementioned six comparable methods on five OHAD datasets. CL-CaGAN is the comparable SOTA method for unsupervised continual OHAD. MAS and EWC are two representative strategies for regular continuous learning method for traditional RGB images, respectively. We change the application form to the OHAD task to verify the effectiveness. Intuitively, these two methods appear to be better suited for the CL scenario. Furthermore, OWM is a representative method of gradient orthogonality in CL family. Therefore, we mainly use these CL-based algorithm to evaluate and demonstrate the effect and performance in this paper.

\textbf{Quantitative Analysis:} As shown in Tables \ref{ACCBWT}, we evaluate the performance of these methods for OHAD tasks, and the experiments cover different tasks combinations from tasks 2 to tasks 5. For instance, tasks 2 means the CL-based OHAD for Los Angeles-1 and Bay Champagne, tasks 5 means the CL-based OHAD from Los Angeles-1 to San Diego. We mainly analysis CL-based strategies and the backbone effect between BioGAN and CaGAN. The results in Table \ref{ACCBWT} indicates that our proposed method CL-BioGAN presents the best $ACC$ for most of OHAD tasks, i.e. tasks 3, tasks 4 and tasks 5. Specifically, this result not only displays the high $ACC$ of CL-BioGAN in processing continuous tasks, but also indicates the satisfying stability and robustness ability to avoid forgetting old tasks when learning new tasks. Parameter comparison further demonstrates that our method achieves high accuracy with fewer parameters while maintaining robust accuracy improvement for learning new tasks.
The testing results of catastrophic forgetting performance on previous tasks are shown in Fig. \ref{AUC-change}, which demonstrated that CL-BioGAN is capable of retaining important knowledge from previous tasks while learning new tasks, thereby presenting more stable AUC for each task. From the $ACC$, $BWT$ and $FWT$ performance presented in Fig. \ref{ACC and BWT}, we can observe that our proposed CL-BioGAN indicates more robustness and reliability for all the tasks 5 in terms of involved transformer structure and biological-inspired AF loss, and exhibits potential practical applications for the proposed CL strategy. Fig. \ref{3DROC} further demonstrates that our method exhibits more stable detection capabilities across different AUC curves. In addition, the background suppression rates are illustrated in Table \ref{BS} as new tasks are learned, while the larger the BS rate means the better discrepancy between anomalies and backgrounds. 

For the HAD100 dataset, we compared the CL-CaGAN and CL-BioGAN methods, as shown in Table \ref{HAD100ACC}. The $ACC$, $BWT$, $FWT$ and $Avg\ BS$ are valued after continual anomaly detection on the five HAD100 datasets. The experiments cover different tasks combinations from tasks 2 to tasks 5. For instance, tasks 2 means the CL-based OHAD for HAD100\_91 and HAD100\_100, tasks 5 means the CL-based OHAD from HAD100\_91 to HAD100\_40. It can be clearly observed that CL-BioGAN achieves a better balance between memory stability and plasticity, resulting in less forgetting. Meanwhile, the Average $BS$ values further reinforcing the conclusion that CL-BioGAN exhibits strong stability in continual anomaly detection.


\begin{figure*}[ht] 
\centering 

\subfigure[]{
\begin{minipage}[t]{0.075 \linewidth}
\centering

\includegraphics[width=15mm,height=15mm]{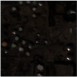}
\vspace{-0.3cm}

\includegraphics[width=15mm,height=15mm]{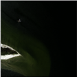}
\vspace{-0.3cm}

\includegraphics[width=15mm,height=15mm]{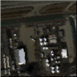}
\vspace{-0.3cm}

\includegraphics[width=15mm,height=15mm]{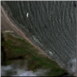}
\vspace{-0.3cm}

\includegraphics[width=15mm,height=15mm]{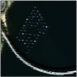}
\vspace{-0.3cm}

\end{minipage}
}
\subfigure[]{
\begin{minipage}[t]{0.075\linewidth}
\centering

\includegraphics[width=15mm,height=15mm]{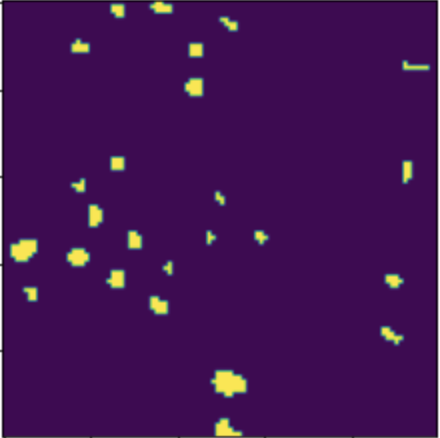}
\vspace{-0.3cm}

\includegraphics[width=15mm,height=15mm]{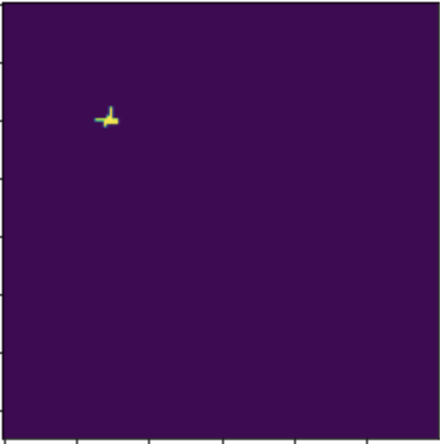}
\vspace{-0.3cm}

\includegraphics[width=15mm,height=15mm]{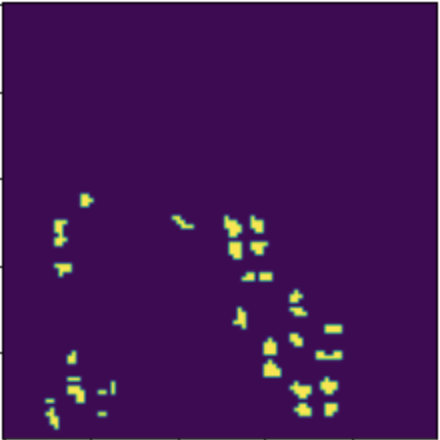}
\vspace{-0.3cm}

\includegraphics[width=15mm,height=15mm]{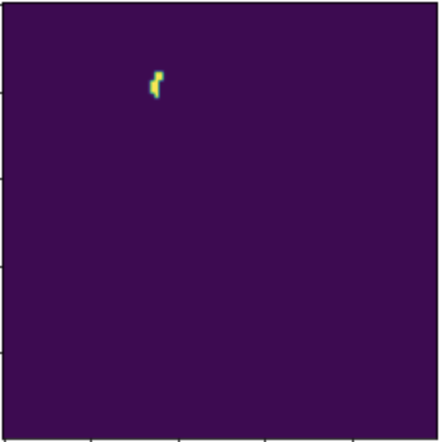}
\vspace{-0.3cm}

\includegraphics[width=15mm,height=15mm]{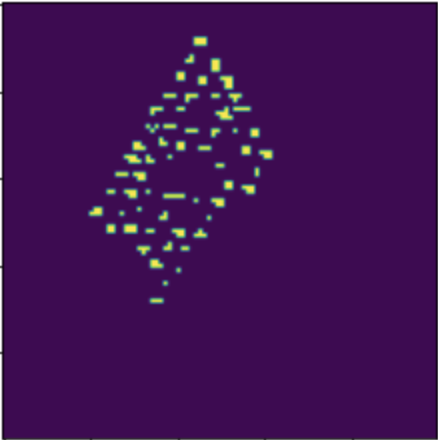}
\vspace{-0.3cm}

\end{minipage}
}
\subfigure[]{
\begin{minipage}[t]{0.075\linewidth}
\centering


\includegraphics[width=15mm,height=15mm]{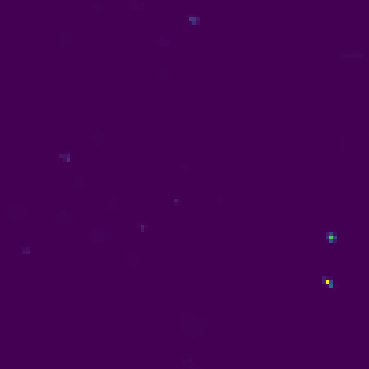}
\vspace{-0.3cm}

\includegraphics[width=15mm,height=15mm]{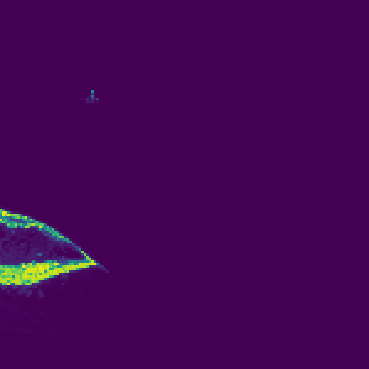}
\vspace{-0.3cm}

\includegraphics[width=15mm,height=15mm]{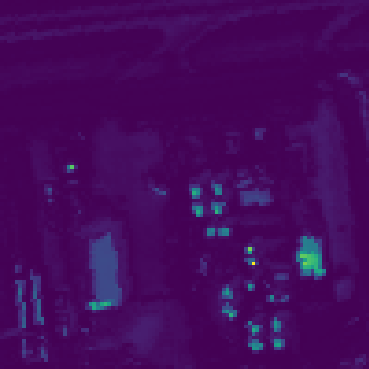}
\vspace{-0.3cm}

\includegraphics[width=15mm,height=15mm]{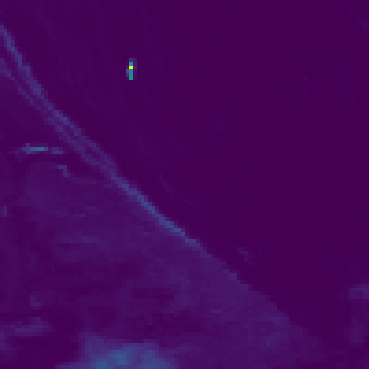}
\vspace{-0.3cm}

\includegraphics[width=15mm,height=15mm]{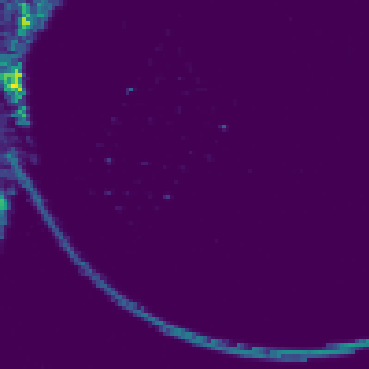}
\vspace{-0.3cm}

\end{minipage}
}
\subfigure[]{
\begin{minipage}[t]{0.075\linewidth}
\centering

\includegraphics[width=15mm,height=15mm]{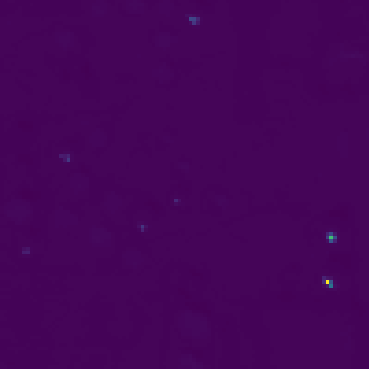}
\vspace{-0.3cm}

\includegraphics[width=15mm,height=15mm]{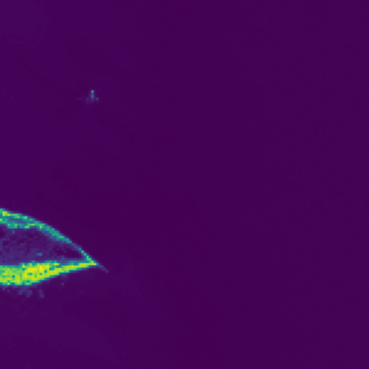}
\vspace{-0.3cm}

\includegraphics[width=15mm,height=15mm]{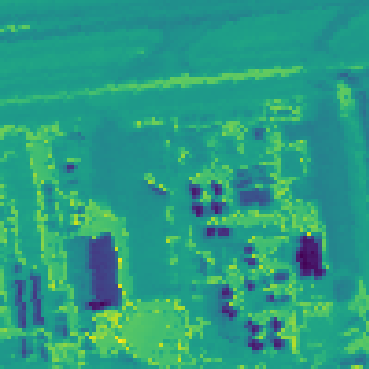}
\vspace{-0.3cm}

\includegraphics[width=15mm,height=15mm]{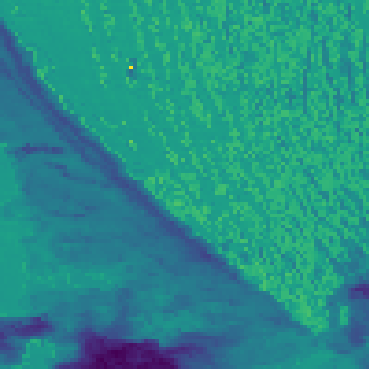}
\vspace{-0.3cm}

\includegraphics[width=15mm,height=15mm]{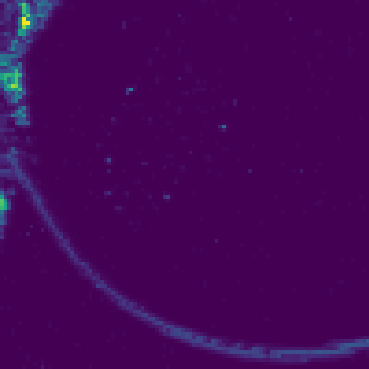}
\vspace{-0.3cm}

\end{minipage}
}
\subfigure[]{
\begin{minipage}[t]{0.075\linewidth}
\centering

\includegraphics[width=15mm,height=15mm]{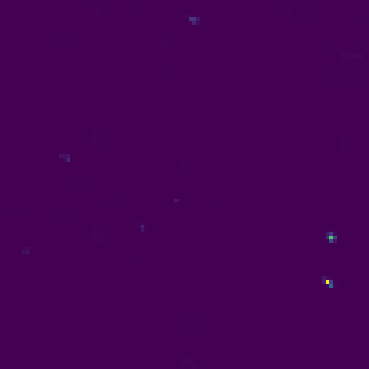}
\vspace{-0.3cm}

\includegraphics[width=15mm,height=15mm]{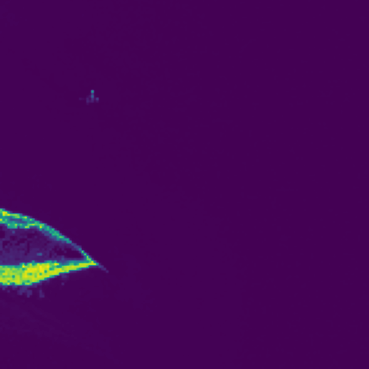}
\vspace{-0.3cm}

\includegraphics[width=15mm,height=15mm]{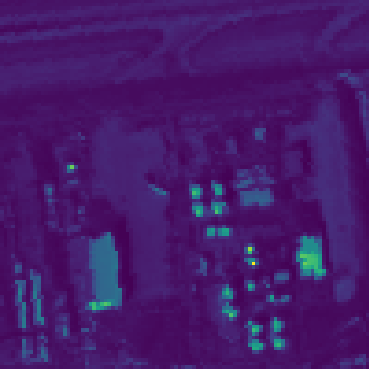}
\vspace{-0.3cm}

\includegraphics[width=15mm,height=15mm]{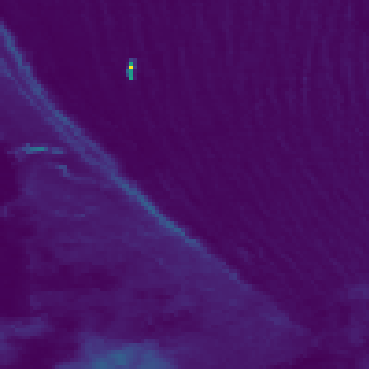}
\vspace{-0.3cm}

\includegraphics[width=15mm,height=15mm]{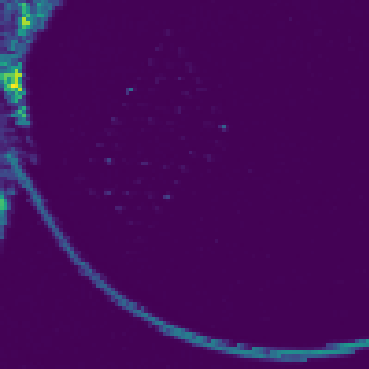}
\vspace{-0.3cm}

\end{minipage}
}
\subfigure[]{
\begin{minipage}[t]{0.075\linewidth}
\centering

\includegraphics[width=15mm,height=15mm]{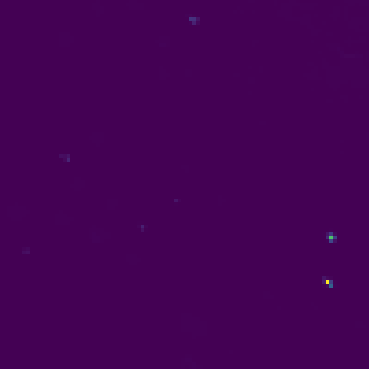}
\vspace{-0.3cm}

\includegraphics[width=15mm,height=15mm]{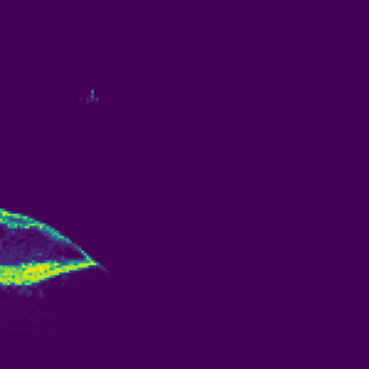}
\vspace{-0.3cm}

\includegraphics[width=15mm,height=15mm]{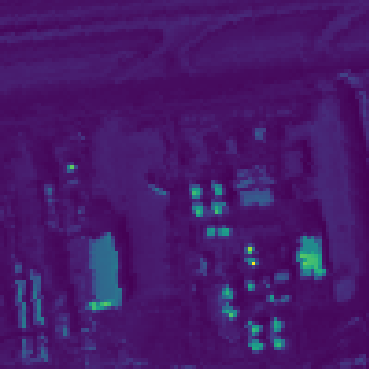}
\vspace{-0.3cm}

\includegraphics[width=15mm,height=15mm]{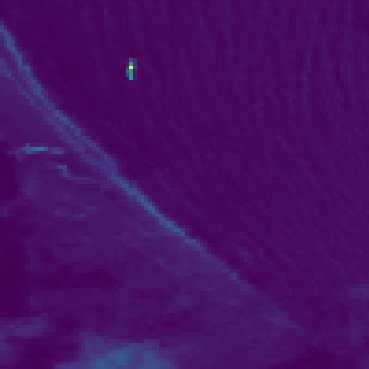}
\vspace{-0.3cm}

\includegraphics[width=15mm,height=15mm]{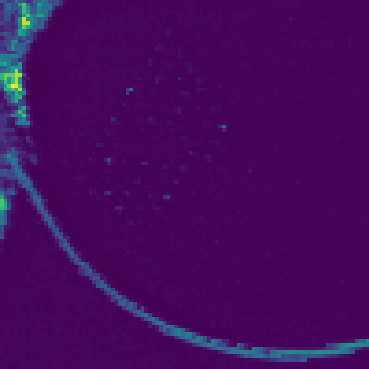}
\vspace{-0.3cm}

\end{minipage}
}
\subfigure[]{
\begin{minipage}[t]{0.075\linewidth}
\centering

\includegraphics[width=15mm,height=15mm]{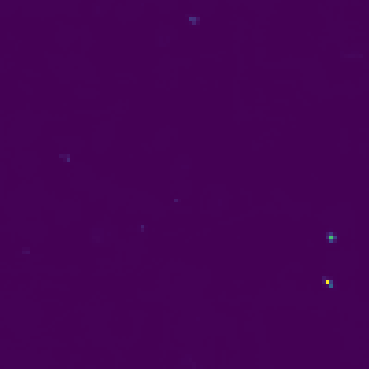}
\vspace{-0.3cm}

\includegraphics[width=15mm,height=15mm]{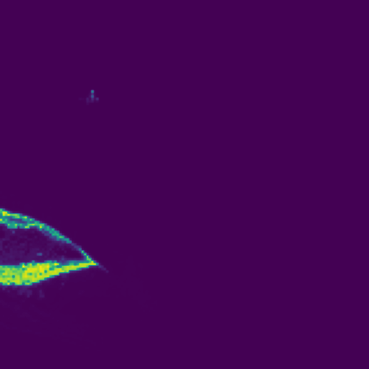}
\vspace{-0.3cm}

\includegraphics[width=15mm,height=15mm]{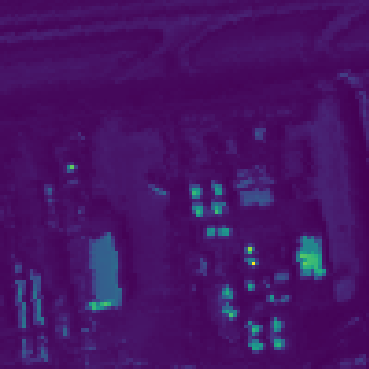}
\vspace{-0.3cm}

\includegraphics[width=15mm,height=15mm]{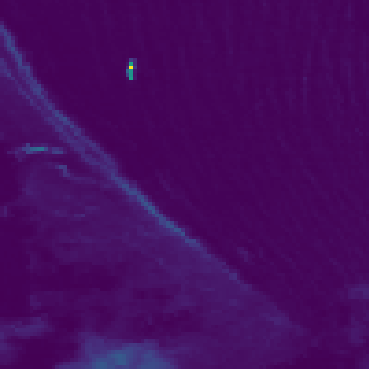}
\vspace{-0.3cm}

\includegraphics[width=15mm,height=15mm]{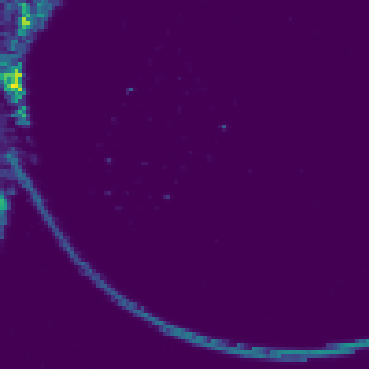}
\vspace{-0.3cm}

\end{minipage}
}
\subfigure[]{
\begin{minipage}[t]{0.075\linewidth}
\centering

\includegraphics[width=15mm,height=15mm]{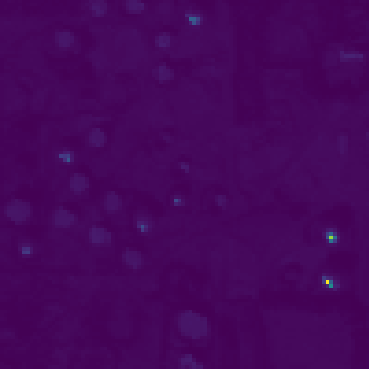}
\vspace{-0.3cm}

\includegraphics[width=15mm,height=15mm]{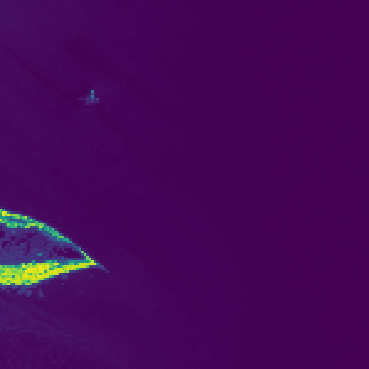}
\vspace{-0.3cm}

\includegraphics[width=15mm,height=15mm]{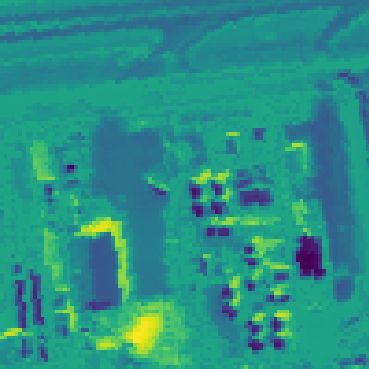}
\vspace{-0.3cm}

\includegraphics[width=15mm,height=15mm]{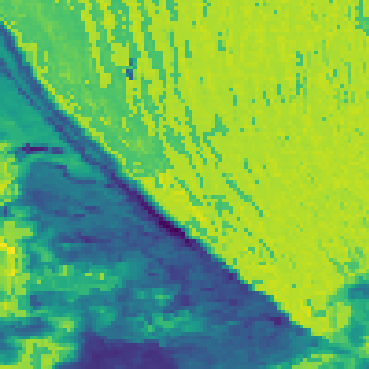}
\vspace{-0.3cm}

\includegraphics[width=15mm,height=15mm]{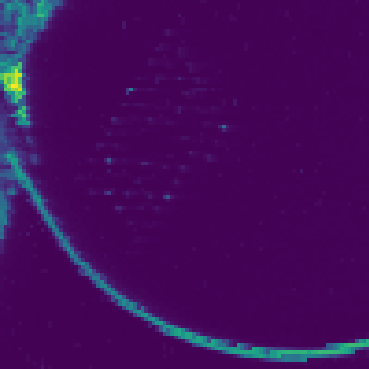}
\vspace{-0.3cm}

\end{minipage}
}
\subfigure[]{
\begin{minipage}[t]{0.065\linewidth}
\centering

\includegraphics[width=15mm,height=15mm]{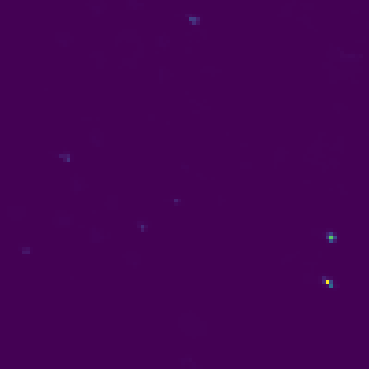}
\vspace{-0.3cm}

\includegraphics[width=15mm,height=15mm]{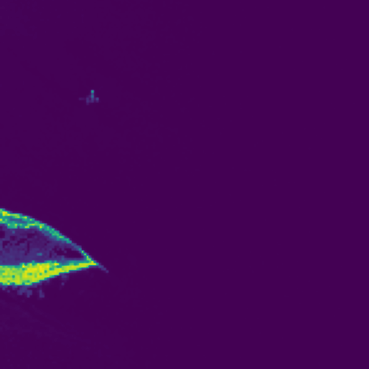}
\vspace{-0.3cm}

\includegraphics[width=15mm,height=15mm]{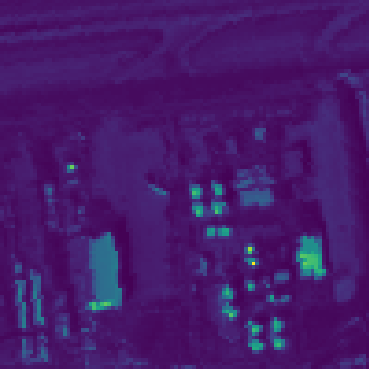}
\vspace{-0.3cm}

\includegraphics[width=15mm,height=15mm]{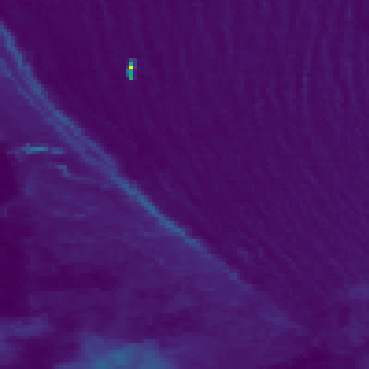}
\vspace{-0.3cm}

\includegraphics[width=15mm,height=15mm]{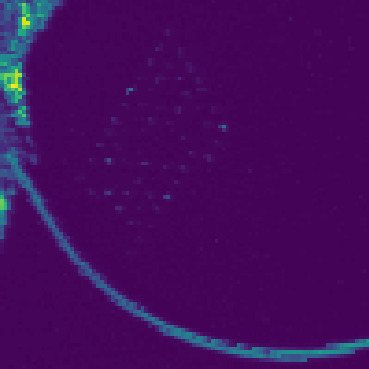}
\vspace{-0.3cm}

\end{minipage}
}
\caption{Visualization for five OHAD results after learning 1-5 Tasks. The data set from top to down is Los Angeles-1, Cat Island, Los Angeles-2, San Diego, Bay Champagne, respectively. (a) False Color image of HAD. (b) Groundtruth map. (c) J-BioGAN. (d) FT-BioGAN. (e) MAS-BioGAN. (f) EWC-BioGAN. (g) OWM-BioGAN (h) CL-CaGAN. (i) CL-BioGAN.} 
\label{detectionBioGAN} 
\end{figure*}
\textbf{Visual Detection Results:} As illustrated in Fig. \ref{detectionBioGAN}, CL-BioGAN demonstrates more stable detect capacity for the previous detection scenes/tasks after learning the knowledge of multiple scenes/tasks, the detection performance  does not affected by surrounding noisy pixels. Compared to CL-CaGAN, we can observe from Fig. \ref{detectionBioGAN} (h) and (i) that our method exhibits more robust detection performance for different anomalies. Meanwhile, CL-BioGAN can minimize the false positives of anomalies in terms of local-global attention mechanism involved in transformer structure with designed AF loss for efficient BS.





\subsection{Ablation Experiment}
\begin{figure}[!ht] 
\centering 
\subfigure[]{
\begin{minipage}[t]{0.5\linewidth}
\centering
\includegraphics[width=40mm,height=35mm]{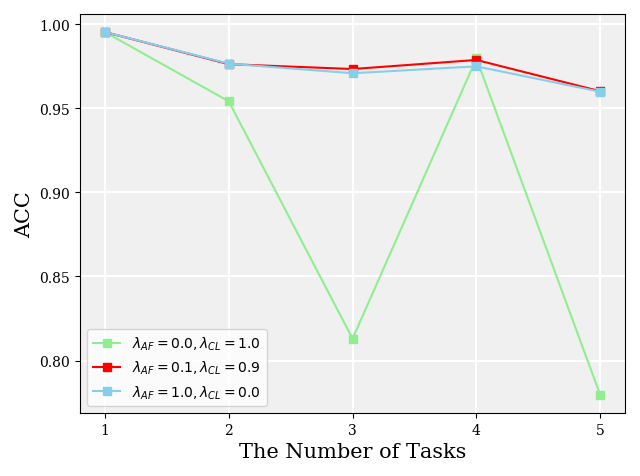}
\end{minipage}
}
\subfigure[]{
\begin{minipage}[t]{0.5\linewidth}
\centering
\includegraphics[width=40mm,height=35mm]{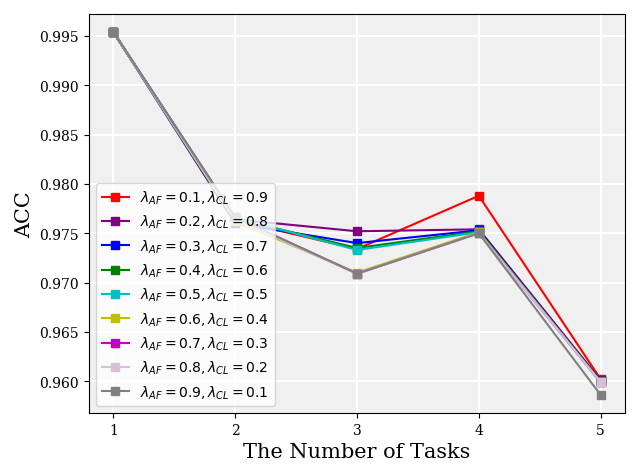}
\end{minipage}
}

\caption{Varied parameter proportions of $\lambda _{AF}$ and $\lambda _{CL}$ for the detection ACC. (a) represents the individual impact of $\lambda _{AF}$ and $\lambda _{CL}$ on the experimental results, while (b) represents the impact of $\lambda _{AF}$ and $\lambda _{CL}$ on the experimental results within a certain range of values.} 
\label{AF} 
\end{figure}

\paragraph{Effect analysis of $\lambda _{AF}$ and $\lambda _{CL}$} A detailed effect and assessment of CL performance for the proportions of $\lambda _{AF}$ and $\lambda _{CL}$ is illustrated in Fig. \ref{AF}. We can observe that the network stability is enhanced with the addition of AF Loss, which helps to learn new knowledge more effectively when new tasks arrive. The results proves that the proposed AF loss and CL loss can efficiently alleviate the issue of catastrophic forgetting in CL. We selected the optimal $\lambda _{AF}$ and $\lambda _{CL}$ value for the following experiments.



\begin{table}[!ht]
\centering
\caption{The result of identical pixels to the total number of pixels(ratio1) and correctly identified background pixels to the actual background pixel count(ratio2) for Euclidean distance(ED) and SAM in round pixels in groundtruth.}
	\label{pixel}
	\setlength{\tabcolsep}{0.5mm}{
	\begin{tabular}{cccccc}
		\toprule  
		\multirow{2}{*}{HSIs} & \multicolumn{2}{c}{ED} &	\multicolumn{2}{c}{SAM}& \multirow{2}{*}{GT}	\\
                 & ratio1 & ratio2 & ratio1 & ratio2 & \\
		\midrule  
  Los Angeles-1 &	0.6832	&0.6798 & \textbf{0.9739}	&\textbf{0.9979} &9728\\
		\midrule  
		Cat Island &0.8782&0.8781& \textbf{0.9893}&\textbf{0.9898}&22481\\
		\midrule  
		Los Angeles-2 &	0.0723&0.0503 & \textbf{0.9762}&	\textbf{0.9991}&9768\\
		\midrule  
		San Diego &	0.0202&0.0196 &\textbf{0.9990}&\textbf{0.9993}&9989\\
		\midrule  
		Bay Champagne &	0.6142&0.6062&\textbf{0.9571}&\textbf{0.9758}&9798\\
		\bottomrule  
	\end{tabular}
	}
\end{table}
\paragraph{Effect analysis for BSM} The comparison between the number of selected pixels and the real background pixels is shown in Table \ref{pixel}. We can notice that background selection based on SAM strategy indicates more accurate background pixels selection in most cases, which also provides the basic information for the generator to accurately learn reconstruction features of the background pixel.
\begin{table}[ht]
	\centering
	\caption{Analyzing the impact of the number of clustering groups for CL-BioGAN. AUC means the value of ${AUC}_{(D,F)}$. The optimal results are highlighted in bold.}
	\setlength{\tabcolsep}{1.4mm}{
	\begin{tabular}{cccccccc}
		\toprule  
ABU Dataset &1& 2 & 3  & 4 & 5 & \multirow{2}{*}{ACC$\uparrow$}&\multirow{2}{*}{BWT$\uparrow$} \\
           $P$& AUC$\uparrow$ & AUC$\uparrow$& AUC$\uparrow$& AUC$\uparrow$& AUC$\uparrow$ \\
        \midrule  
$P=2$ & 0.9954 & - & - & -& - & 0.9954 & -\\
$P=3$ & 0.9954 & - & - & -& - & 0.9954 & -\\
$P=4$ & 0.9954 & - & - & -& - & 0.9954 & -\\
$P=5$ & 0.9954 & - & - & -& - & 0.9954 & -\\
$P=6$ & 0.9954 & - & - & -& - & 0.9954 & -\\
         \midrule  
$P=2$ & 0.9629	&\textbf{0.9713} & - & -& - & 0.9671 & -0.0325\\
$P=3$ & \textbf{0.9821}	&0.9704 & - & -& - & \textbf{0.9763}&\textbf{-0.0133}\\
$P=4$ & 0.9635	&\textbf{0.9713} & - & -& - & 0.9674 & -0.0319\\
$P=5$ & 0.9632	&\textbf{0.9713}& - & -& - & 0.9673& -0.0322\\
$P=6$ & 0.9640	&0.9712& - & -& - & 0.9676 & -0.0314\\
         \midrule  
$P=2$ & 0.9708	&\textbf{0.9716}	&0.9684 & -& - & 0.9703&\textbf{0.0041}\\
$P=3$ & \textbf{0.9817}	&0.9713	&0.9672& -& - & 0.9734&0.0003\\
$P=4$ & 0.9709	&\textbf{0.9716}	&0.9684& -& - & 0.9703&0.0039\\
$P=5$ & 0.9707	&\textbf{0.9716}	&0.9684& -& - & 0.9702&0.0039\\
$P=6$ & 0.9664	&0.9708	&\textbf{0.9835}& -& - & \textbf{0.9736}&0.0010\\
         \midrule  
$P=2$ & 0.9908	&\textbf{0.9710}	&\textbf{0.9658}	&\textbf{0.9876}& - &\textbf{0.9788}&\textbf{0.0056}\\
$P=3$ & \textbf{0.9911}	&\textbf{0.9710}&0.9656	&0.9875& - & \textbf{0.9788} &0.0025\\
$P=4$ & 0.9907	&\textbf{0.9710}	&0.9656	&0.9875& - & 0.9787 &0.0055\\
$P=5$ & 0.9904	&0.9709	&0.9656	&0.9875& - &0.9786 &0.0054\\
$P=6$ & 0.9880	&0.9709	&0.9656&\textbf{0.9876}& - & 0.9780 &0.0013\\
         \midrule  
$P=2$ & 0.9753&0.9705&	\textbf{0.9659}&	\textbf{0.9876}&	0.9005&	0.9600&-0.0040\\
$P=3$ &0.9786 &0.9701&\textbf{0.9659}&0.9874&	0.8992&	0.9602&-0.0033\\
$P=4$ &0.9742&\textbf{0.9706}&0.9658&\textbf{0.9876}&\textbf{0.9007}&0.9598&-0.0042\\
$P=5$ & \textbf{0.9865}&0.9702&0.9656&0.9874&0.8995&\textbf{0.9618}&\textbf{-0.0012}\\
$P=6$ &0.9805&0.9702&\textbf{0.9659}&0.9874&	0.8994&	0.9607&-0.0020\\
		\bottomrule  
	\end{tabular}
	}	

	\label{P}
\end{table}
\paragraph{Effect analysis of the number of clustering group} Table \ref{P} provides an insightful analysis of the impact of varying cluster numbers $P$ in the K-Means algorithm for CL OHAD task performance within the sample replay strategy framework. Across the range of cluster numbers studied from $P$=2 to $P$=6, we can notice that the configuration with $P$=3 consistently yields the highest accuracy.
While configurations with $P$=5 and $P$=6 indicates slightly increasing $BWT$ in certain tasks, $P$=3 provides a more consistent performance across all task configurations. Therefore, the choice of $P$=3 represents an optimal balance between $ACC$ and $BWT$. Although other $P$ values exhibit minor advantages in individual metrics, $P$=3 delivers the optimal or sub-optimal performance across all principal indicators.

\begin{table}[ht]
	\centering
	\caption{The ACC and BWT impact analyses for different components of proposed CL-BioGAN. w/o R means exclude of Replay, w/o T means exclude of $\textbf L_{2}$ SA, w/o $L_{AF}$ means exclude of AF Loss, w/o $L_{CL}$ means exclude of CL Loss and ours means CL-BioGAN. AUC represent the value of ${AUC}_{(D,F)}$. The optimal results are highlighted in bold.}
	\setlength{\tabcolsep}{1.4mm}{
	\begin{tabular}{cccccccc}
		\toprule  
ABU Tasks &1& 2 & 3  & 4 & 5 & \multirow{2}{*}{ACC$\uparrow$}&\multirow{2}{*}{BWT$\uparrow$} \\
           method& AUC$\uparrow$ & AUC$\uparrow$& AUC$\uparrow$& AUC$\uparrow$& AUC$\uparrow$ \\
        \midrule  
        w/o R & 0.9954 & - & - & -& - & 0.9954 & -\\
        w/o T & \textbf{0.9963} & - & - & -& - & \textbf{0.9963} & -\\
	w/o $L_{AF}$ & 0.9954 & - & - & -& - & 0.9954 & -\\
    w/o $L_{CL}$ & 0.9954 & - & - & -& - & 0.9954 & -\\
        ours & 0.9954 & - & - & -& - & 0.9954 & -\\
         \midrule  
w/o R & 0.9827&0.9703 & - & -& - & 0.9765 & -0.0127\\
w/o T & 0.9834 &0.9693 & - & -& - & 0.9764&-0.0129\\
w/o $L_{AF}$ & 0.9830 & 0.9702 & - & -& - & \textbf{0.9766} & -0.0124\\
w/o $L_{CL}$ & \textbf{0.9884} & 0.9200 & - & -& - & 0.9542 & \textbf{-0.0070}\\
ours & 0.9821&\textbf{0.9704} & - & -& - & 0.9763 & -0.0133\\
         \midrule  
w/o R & 0.9767& 0.9700& 0.9659 & -& - & 0.9709& -0.0032\\
w/o T & 0.9814	&0.9695	&0.9658 & -& - & 0.9722&-0.0009\\
w/o $L_{AF}$& 0.9767& 0.9700& 0.9659 & -& - & 0.9709& -0.0033\\
w/o $L_{CL}$& 0.5594& 0.9207& 0.9590 & -& - & 0.8130& -0.2142\\
ours & \textbf{0.9817}	&\textbf{0.9713}&\textbf{0.9672}& -& - & \textbf{0.9734} &\textbf{0.0003}\\
         \midrule  
w/o R & 0.9767& 0.9700& 0.9659	&0.9874& - &0.9750&	0.0000\\
w/o T & 0.9804	&0.9700&0.9658	&0.9873& - & 0.9759 &-0.0002\\
w/o $L_{AF}$& 0.9767& 0.9700& 0.9659	&0.9874& - &0.9750&	0.0000\\
w/o $L_{CL}$& \textbf{0.9961}& 0.9685& \textbf{0.9668}& 0.9874& - &\textbf{0.9797}& \textbf{0.1641}\\
ours & 0.9911&	\textbf{0.9710}&	0.9656	&\textbf{0.9875}& - & 0.9788	&0.0025\\
         \midrule  
w/o R & 0.9767& 0.9700& 0.9659	&0.9874& \textbf{0.8998} & 0.9599 & \textbf{0.0000}\\
w/o T & 0.9256	&\textbf{0.9705}&\textbf{0.9669}&	\textbf{0.9875}&	0.8973& 0.9496 & -0.0133\\
w/o $L_{AF}$& 0.9767& 0.9700& 0.9659& 0.9874& \textbf{0.8998} & 0.9599 &	\textbf{0.0000}\\
w/o $L_{CL}$& 0.9331& 0.1149& 0.9676& \textbf{0.9875}& 0.8944 & 0.7795 &-0.2289\\
ours & \textbf{0.9786}& 0.9701& 0.9659 & 0.9874 &0.8992& \textbf{0.9602} &	-0.0033\\

		\bottomrule  
	\end{tabular}
	}	

	\label{ablation}
\end{table}

\begin{table}[ht]
	\centering
	\caption{The ${AUC}_{BS}$ impact analyses for different components of CL-BioGAN. The optimal results are highlighted in bold.}
	\setlength{\tabcolsep}{1.4mm}{
	\begin{tabular}{ccccccc}
		\toprule  
  ABU Tasks &1& 2 & 3  & 4 & 5 & \multirow{2}{*}{Avg$\uparrow$}\\
           Methods& ${BS}$$\uparrow$ & ${BS}$$\uparrow$& ${BS}$$\uparrow$& ${BS}$$\uparrow$& ${BS}$$\uparrow$ \\
          \midrule  
        w/o R & \textbf{0.9952}  & - & - & -& - & \textbf{0.9952} \\
        w/o T & 0.9951 & - & - & -& - & 0.9951 \\
	w/o $L_{AF}$ & \textbf{0.9952} & - & - & -& - & \textbf{0.9952} \\
        w/o $L_{CL}$ &\textbf{0.9952} & - & - & -& - & \textbf{0.9952} \\
        ours & \textbf{0.9952}  & - & - & -& - & \textbf{0.9952} \\
         \midrule  
w/o R & 0.9808& 0.9526 & - & -& - & 0.9667 \\
w/o T& 0.9820 &0.9520 & - & -& - & 0.9670\\
w/o $L_{AF}$& 0.9810 & \textbf{0.9545} & - & -& - & \textbf{0.9697} \\
w/o $L_{CL}$& \textbf{0.9830} & 0.8906 & - & -& - & 0.9368 \\
ours& 0.9805&0.9530 & - & -& - & 0.9668 \\
         \midrule  
w/o R & 0.9744& 0.9523& 0.8933 & -& - & 0.9400\\
w/o T & 0.9800	&0.9526	&0.8970 & -& - & 0.9432\\
w/o $L_{AF}$& 0.9744& 0.9523& 0.8933& -& - & 0.9400\\
w/o $L_{CL}$& 0.5553& 0.8987& 0.8762& -& - & 0.7767\\
ours& \textbf{0.9803}	&\textbf{0.9544}&\textbf{0.9018}& -& - & \textbf{0.9455}\\
         \midrule  
w/o R & 0.9743	&0.9522	&0.8933	&0.9295& - & 0.9373\\
w/o T & 0.9785	&0.9528&0.8955	&0.9318& - & 0.9397\\
w/o $L_{AF}$& 0.9743&0.9522 &0.8933 &0.9295& - & 0.9373 \\
w/o $L_{CL}$& \textbf{0.9954}&\textbf{0.9541}&	\textbf{0.9135}&	\textbf{0.9511}& - &\textbf{0.9535}\\
ours& 0.9889&	0.9534&0.8943	&0.9328& - & 0.9424	\\
         \midrule  
w/o R & 0.9742&0.9522&0.8932 &0.9294 &0.8620&	0.9222\\
w/o T & 0.9231	&0.9520&0.8949&	0.9163&0.8564&	0.9085\\
w/o $L_{AF}$& 0.9742&0.9522&0.8932&0.9294 &0.8620 &0.9222	\\
w/o $L_{CL}$& 0.9287&-0.0023& \textbf{0.9076} &\textbf{0.9421} &\textbf{0.8678}& 0.7288\\
ours&\textbf{0.9764}&\textbf{0.9526}&0.8942&0.9304&0.8625& \textbf{0.9232}\\
		\bottomrule  
	\end{tabular}
	}	

	\label{BS_T}
\end{table}

\paragraph{Other ablation effect experiment}
We mainly evaluate and analyze the impact of the $\textbf L_{2}$ SA structure in $D$, the Replay strategy, the AF loss and the CL loss for OHAD performance across five different scenario tasks. The effect for above mentioned components utilized in our proposed CL-BioGAN are illustrated in Table \ref{ablation}, respectively. We can observe that CL-BioGAN with all above mentioned components presents the optimal $ACC$ and $BWT$ results in most cases especially after CL from the second task. For the modal of CL-BioGAN without AF Loss, it exhibits unstable changes in $ACC$ and significant fluctuations in $BWT$ when learning new scenario tasks. In contrast, CL-BioGAN demonstrates more stable learning of new tasks while effectively preventing catastrophic forgetting issues. Particularly, after learning the third scenario, the CL-BioGAN demonstrates beneficial retention of knowledge for previous scenarios, which indicates that AF Loss enhances sustained detection performance while increasing the robustness and stability of the modal.

Table \ref{BS_T} reflects the ${AUC}_{BS}$ impact analyses for different components of the proposed CL-BioGAN. As new tasks are learned, AF Loss may degrade the background suppression performance in certain scenarios, while AF Loss ensures the stability of background suppression. While for the CL-BioGAN
Without AF Loss, it exhibits significant fluctuations as new knowledge is incorporated. This performance indicates that AF Loss is a key component for sustaining CL ability and served as a stabilizer for BS. The similar results between ${AUC}_{(D,F)}$ in Table \ref{ablation} and ${AUC}_{BS}$ in Table \ref{BS_T} indicate that CL-BioGAN reduces false positives in ${AUC}_{(D,F)}$ as well as enhances the ${AUC}_{BS}$ performance, which further demonstrates the proposed CL-BioGAN can facilitate better generation for pseudo background pixels in OHAD.

\section{Conclusion}
In this study, in order to mitigate catastrophic forgetting as well as adapt to new endless coming tasks in the cross-domain OHAD, a novel CL-BioGAN is proposed to improve the detection accuracy and robustness through a designed Bio-inspired AF loss and self-attention-based GAN structure. The proposed CL-BioGAN incorporates a continuous replay strategy by CL-Bio loss to preserve historical knowledge and adapt to new tasks in open scenario circumstance. AF loss is designed to balance stability and plasticity by modulating the forgetting rate of previous knowledge, which is functionally consistent with the advantage of biological forgetting in memory flexibility. This connection provides a underlying mechanism theoretical explanation for the conflict between biological active forgetting and continually learning experiences. In addition, CL-BioGAN achieves an end-to-end reconstruction by the synergistic integration of a generator and an $L_2$ SA discriminator within a GAN framework for effectively learning representative spectral characteristics of background distribution through differentiable data augmentation. Our experiments on five cross-domain OHAD datasets indicate that CL-BioGAN significantly offer more stable OHAD performance while maintaining a balance between historical and new tasks.


\bibliographystyle{IEEEtran}  
\bibliography{main}

\begin{thebibliography}{10}
\providecommand{\url}[1]{#1}
\csname url@samestyle\endcsname
\providecommand{\newblock}{\relax}
\providecommand{\bibinfo}[2]{#2}
\providecommand{\BIBentrySTDinterwordspacing}{\spaceskip=0pt\relax}
\providecommand{\BIBentryALTinterwordstretchfactor}{4}
\providecommand{\BIBentryALTinterwordspacing}{\spaceskip=\fontdimen2\font plus
\BIBentryALTinterwordstretchfactor\fontdimen3\font minus
  \fontdimen4\font\relax}
\providecommand{\BIBforeignlanguage}[2]{{%
\expandafter\ifx\csname l@#1\endcsname\relax
\typeout{** WARNING: IEEEtran.bst: No hyphenation pattern has been}%
\typeout{** loaded for the language `#1'. Using the pattern for}%
\typeout{** the default language instead.}%
\else
\language=\csname l@#1\endcsname
\fi
#2}}
\providecommand{\BIBdecl}{\relax}
\BIBdecl

\bibitem{77}
Y.~Duan, C.~Chen, M.~Fu, Y.~Li, X.~Gong, and F.~Luo, ``Dimensionality reduction
  via multiple neighborhood-aware nonlinear collaborative analysis for
  hyperspectral image classification,'' \emph{IEEE Transactions on Circuits and
  Systems for Video Technology}, pp. 1--1, 2024.

\bibitem{-1}
Z.~Li, Y.~Zhang, and J.~Zhang, ``Hyperspectral anomaly detection for spectral
  anomaly targets via spatial and spectral constraints,'' \emph{IEEE
  Transactions on Geoscience and Remote Sensing}, vol.~60, pp. 1--15, 2022.

\bibitem{0}
J.~Li, X.~Wang, S.~Wang, H.~Zhao, and Y.~Zhong, ``One-step detection paradigm
  for hyperspectral anomaly detection via spectral deviation relationship
  learning,'' \emph{IEEE Transactions on Geoscience and Remote Sensing},
  vol.~62, pp. 1--15, 2024.

\bibitem{78}
L.~Song, Z.~Feng, S.~Yang, X.~Zhang, and L.~Jiao, ``Interactive
  spectral-spatial transformer for hyperspectral image classification,''
  \emph{IEEE Transactions on Circuits and Systems for Video Technology}, pp.
  1--1, 2024.

\bibitem{1}
G.~I. Parisi, R.~Kemker, J.~L. Part, C.~Kanan, and S.~Wermter, ``Continual
  lifelong learning with neural networks: A review,'' \emph{Neural Networks},
  vol. 113, pp. 54--71, 2019.

\bibitem{2}
M.~De~Lange, R.~Aljundi, M.~Masana, S.~Parisot, X.~Jia, A.~Leonardis,
  G.~Slabaugh, and T.~Tuytelaars, ``A continual learning survey: Defying
  forgetting in classification tasks,'' \emph{IEEE Transactions on Pattern
  Analysis and Machine Intelligence}, vol.~44, no.~7, pp. 3366--3385, 2022.

\bibitem{87}
M.~Masana, X.~Liu, B.~Twardowski, M.~Menta, A.~D. Bagdanov, and J.~van~de
  Weijer, ``Class-incremental learning: Survey and performance evaluation on
  image classification,'' \emph{IEEE Transactions on Pattern Analysis and
  Machine Intelligence}, vol.~45, no.~5, pp. 5513--5533, 2023.

\bibitem{4}
M.~McCloskey and N.~J. Cohen, ``Catastrophic interference in connectionist
  networks: The sequential learning problem,'' ser. Psychology of Learning and
  Motivation.\hskip 1em plus 0.5em minus 0.4em\relax Academic Press, 1989,
  vol.~24, pp. 109--165.

\bibitem{88}
S.~Dohare, J.~F. Hernandez-Garcia, Q.~Lan, P.~Rahman, A.~R. Mahmood, and R.~S.
  Sutton, ``Loss of plasticity in deep continual learning,'' \emph{Nature},
  vol. 632, no. 8026, pp. 768--774, 2024.

\bibitem{-7}
L.~Wang, X.~Zhang, H.~Su, and J.~Zhu, ``A comprehensive survey of continual
  learning: Theory, method and application,'' \emph{IEEE Transactions on
  Pattern Analysis and Machine Intelligence}, pp. 1--20, 2024.

\bibitem{79}
K.~Li, H.~Chen, J.~Wan, and S.~Yu, ``Esdb: Expand the shrinking decision
  boundary via one-to-many information matching for continual learning with
  small memory,'' \emph{IEEE Transactions on Circuits and Systems for Video
  Technology}, pp. 1--1, 2024.

\bibitem{wang2023incorporating}
L.~Wang, X.~Zhang, Q.~Li, M.~Zhang, H.~Su, J.~Zhu, and Y.~Zhong,
  ``Incorporating neuro-inspired adaptability for continual learning in
  artificial intelligence,'' \emph{Nature Machine Intelligence}, vol.~5,
  no.~12, pp. 1356--1368, 2023.

\bibitem{6}
L.~Wang, M.~Zhang, Z.~Jia, Q.~Li, C.~Bao, K.~Ma, J.~Zhu, and Y.~Zhong,
  ``{AFEC}: Active forgetting of negative transfer in continual learning,'' in
  \emph{Advances in Neural Information Processing Systems}, M.~Ranzato,
  A.~Beygelzimer, Y.~Dauphin, P.~Liang, and J.~W. Vaughan, Eds., vol.~34.\hskip
  1em plus 0.5em minus 0.4em\relax Curran Associates, Inc., 2021, pp.
  22\,379--22\,391.

\bibitem{7}
L.~Aitchison, J.~Jegminat, J.~A. Menendez, J.-P. Pfister, A.~Pouget, and P.~E.
  Latham, ``Synaptic plasticity as bayesian inference,'' \emph{Nature
  neuroscience}, vol.~24, no.~4, pp. 565--571, 2021.

\bibitem{8}
S.~Schug, F.~Benzing, and A.~Steger, ``Presynaptic stochasticity improves
  energy efficiency and helps alleviate the stability-plasticity dilemma,''
  \emph{eLife}, vol.~10, p. e69884, oct 2021.

\bibitem{9}
R.~Hadsell, D.~Rao, A.~A. Rusu, and R.~Pascanu, ``Embracing change: Continual
  learning in deep neural networks,'' \emph{Trends in cognitive sciences},
  vol.~24, no.~12, pp. 1028--1040, 2020.

\bibitem{14-2}
J.~Kirkpatrick, R.~Pascanu, N.~Rabinowitz, J.~Veness, G.~Desjardins, A.~A.
  Rusu, K.~Milan, J.~Quan, T.~Ramalho, A.~Grabska-Barwinska, D.~Hassabis,
  C.~Clopath, D.~Kumaran, and R.~Hadsell, ``Overcoming catastrophic forgetting
  in neural networks,'' \emph{Proceedings of the National Academy of Sciences},
  vol. 114, no.~13, pp. 3521--3526, 2017.

\bibitem{14-3}
C.~I. Tang, L.~Qendro, D.~Spathis, F.~Kawsar, C.~Mascolo, and A.~Mathur,
  ``Kaizen: Practical self-supervised continual learning with continual
  fine-tuning,'' in \emph{Proceedings of the IEEE/CVF Winter Conference on
  Applications of Computer Vision (WACV)}, January 2024, pp. 2841--2850.

\bibitem{14-1}
Y.~Zhang, L.~Charlin, R.~Zemel, and M.~Ren, ``Integrating present and past in
  unsupervised continual learning,'' \emph{arXiv preprint arXiv:2404.19132},
  2024.

\bibitem{11}
Z.~Huang, Z.~Chen, Z.~Chen, E.~Zhou, X.~Xu, R.~S.~M. Goh, Y.~Liu, C.~Feng, and
  W.~Zuo, ``Learning prompt with distribution-based feature replay for few-shot
  class-incremental learning,'' \emph{arXiv preprint arXiv:2401.01598}, 2024.

\bibitem{12}
X.~Li, H.~Li, and L.~Ma, ``Continual learning of medical image classification
  based on feature replay,'' in \emph{2022 16th IEEE International Conference
  on Signal Processing (ICSP)}, vol.~1, 2022, pp. 426--430.

\bibitem{13}
H.~Cheng, H.~Wen, X.~Zhang, H.~Qiu, L.~Wang, and H.~Li, ``Contrastive
  continuity on augmentation stability rehearsal for continual self-supervised
  learning,'' in \emph{Proceedings of the IEEE/CVF International Conference on
  Computer Vision (ICCV)}, October 2023, pp. 5707--5717.

\bibitem{14}
K.~Jeeveswaran, P.~Bhat, B.~Zonooz, and E.~Arani, ``{BiRT}: Bio-inspired replay
  in vision transformers for continual learning,'' \emph{arXiv preprint
  arXiv:2305.04769}, 2023.

\bibitem{90}
S.~Chen, M.~Zhang, J.~Zhang, and K.~Huang, ``Exemplar-based continual learning
  via contrastive learning,'' \emph{IEEE Transactions on Artificial
  Intelligence}, vol.~5, no.~7, pp. 3313--3324, 2024.

\bibitem{16-1}
Q.~Wang, R.~Wang, Y.~Li, D.~Wei, H.~Wang, K.~Ma, Y.~Zheng, and D.~Meng,
  ``Relational experience replay: Continual learning by adaptively tuning
  task-wise relationship,'' \emph{IEEE Transactions on Multimedia}, pp. 1--15,
  2024.

\bibitem{17}
J.~Zhang, J.~Zhang, S.~Ghosh, D.~Li, S.~Tasci, L.~Heck, H.~Zhang, and C.-C.~J.
  Kuo, ``Class-incremental learning via deep model consolidation,'' in
  \emph{Proceedings of the IEEE/CVF Winter Conference on Applications of
  Computer Vision (WACV)}, March 2020, pp. 1131--1140.

\bibitem{18}
M.~H. Phan, T.-A. Ta, S.~L. Phung, L.~Tran-Thanh, and A.~Bouzerdoum, ``Class
  similarity weighted knowledge distillation for continual semantic
  segmentation,'' in \emph{Proceedings of the IEEE/CVF Conference on Computer
  Vision and Pattern Recognition (CVPR)}, June 2022, pp. 16\,866--16\,875.

\bibitem{19}
A.~S. Alakooz and N.~Ammour, ``A contrastive continual learning for the
  classification of remote sensing imagery,'' in \emph{IGARSS 2022 - 2022 IEEE
  International Geoscience and Remote Sensing Symposium}, 2022, pp. 7902--7905.

\bibitem{20}
V.~Marsocci and S.~Scardapane, ``Continual barlow twins: Continual
  self-supervised learning for remote sensing semantic segmentation,''
  \emph{IEEE Journal of Selected Topics in Applied Earth Observations and
  Remote Sensing}, vol.~16, pp. 5049--5060, 2023.

\bibitem{21}
S.~Kim, L.~Noci, A.~Orvieto, and T.~Hofmann, ``Achieving a better
  stability-plasticity trade-off via auxiliary networks in continual
  learning,'' in \emph{Proceedings of the IEEE/CVF Conference on Computer
  Vision and Pattern Recognition (CVPR)}, June 2023, pp. 11\,930--11\,939.

\bibitem{22}
A.~Mallya, D.~Davis, and S.~Lazebnik, ``Piggyback: Adapting a single network to
  multiple tasks by learning to mask weights,'' in \emph{Proceedings of the
  European Conference on Computer Vision (ECCV)}, September 2018, pp. 67--82.

\bibitem{23}
X.~Li, Y.~Zhou, T.~Wu, R.~Socher, and C.~Xiong, ``Learn to grow: A continual
  structure learning framework for overcoming catastrophic forgetting,'' in
  \emph{Proceedings of the 36th International Conference on Machine Learning},
  ser. Proceedings of Machine Learning Research, K.~Chaudhuri and
  R.~Salakhutdinov, Eds., vol.~97.\hskip 1em plus 0.5em minus 0.4em\relax PMLR,
  09--15 Jun 2019, pp. 3925--3934.

\bibitem{24}
M.~Xue, H.~Zhang, J.~Song, and M.~Song, ``{Meta-Attention} for {ViT}-backed
  continual learning,'' in \emph{Proceedings of the IEEE/CVF Conference on
  Computer Vision and Pattern Recognition (CVPR)}, June 2022, pp. 150--159.

\bibitem{25}
S.~Ebrahimi, F.~Meier, R.~Calandra, T.~Darrell, and M.~Rohrbach, ``Adversarial
  continual learning,'' in \emph{Computer Vision -- ECCV 2020}, A.~Vedaldi,
  H.~Bischof, T.~Brox, and J.-M. Frahm, Eds.\hskip 1em plus 0.5em minus
  0.4em\relax Cham: Springer International Publishing, 2020, pp. 386--402.

\bibitem{86}
H.~E, Y.~Cui, W.~Pedrycz, and Z.~Li, ``Design of distributed rule-based models
  in the presence of large data,'' \emph{IEEE Transactions on Fuzzy Systems},
  vol.~31, no.~7, pp. 2479--2486, 2023.

\bibitem{92}
S.~Wang, X.~Li, J.~Sun, and Z.~Xu, ``Training networks in null space of feature
  covariance for continual learning,'' in \emph{Proceedings of the IEEE/CVF
  conference on Computer Vision and Pattern Recognition}, 2021, pp. 184--193.

\bibitem{93}
Y.~Kong, L.~Liu, Z.~Wang, and D.~Tao, ``Balancing stability and plasticity
  through advanced null space in continual learning,'' in \emph{European
  Conference on Computer Vision}.\hskip 1em plus 0.5em minus 0.4em\relax
  Springer, 2022, pp. 219--236.

\bibitem{94}
R.~Hadsell, D.~Rao, A.~A. Rusu, and R.~Pascanu, ``Embracing change: Continual
  learning in deep neural networks,'' \emph{Trends in cognitive sciences},
  vol.~24, no.~12, pp. 1028--1040, 2020.

\bibitem{95}
R.~Wang, Y.~Bao, B.~Zhang, J.~Liu, W.~Zhu, and G.~Guo, ``Anti-retroactive
  interference for lifelong learning,'' in \emph{European Conference on
  Computer Vision}.\hskip 1em plus 0.5em minus 0.4em\relax Springer, 2022, pp.
  163--178.

\bibitem{96}
J.~Hurtado, A.~Raymond, and A.~Soto, ``Optimizing reusable knowledge for
  continual learning via metalearning,'' \emph{Advances in Neural Information
  Processing Systems}, vol.~34, pp. 14\,150--14\,162, 2021.

\bibitem{97}
G.~Lin, H.~Chu, and H.~Lai, ``Towards better plasticity-stability trade-off in
  incremental learning: A simple linear connector,'' in \emph{Proceedings of
  the IEEE/CVF Conference on Computer Vision and Pattern Recognition}, 2022,
  pp. 89--98.

\bibitem{98}
S.~Wang, X.~Li, J.~Sun, and Z.~Xu, ``Training networks in null space of feature
  covariance for continual learning,'' in \emph{Proceedings of the IEEE/CVF
  conference on Computer Vision and Pattern Recognition}, 2021, pp. 184--193.

\bibitem{99}
S.~Purushwalkam, P.~Morgado, and A.~Gupta, ``The challenges of continuous
  self-supervised learning,'' in \emph{European Conference on Computer
  Vision}.\hskip 1em plus 0.5em minus 0.4em\relax Springer, 2022, pp. 702--721.

\bibitem{100}
E.~Fini, V.~G.~T. Da~Costa, X.~Alameda-Pineda, E.~Ricci, K.~Alahari, and
  J.~Mairal, ``Self-supervised models are continual learners,'' in
  \emph{Proceedings of the IEEE/CVF Conference on Computer Vision and Pattern
  Recognition}, 2022, pp. 9621--9630.

\bibitem{101}
Z.~Wang, Z.~Zhang, C.-Y. Lee, H.~Zhang, R.~Sun, X.~Ren, G.~Su, V.~Perot, J.~Dy,
  and T.~Pfister, ``Learning to prompt for continual learning,'' in
  \emph{Proceedings of the IEEE/CVF conference on computer vision and pattern
  recognition}, 2022, pp. 139--149.

\bibitem{102}
J.~S. Smith, L.~Karlinsky, V.~Gutta, P.~Cascante-Bonilla, D.~Kim, A.~Arbelle,
  R.~Panda, R.~Feris, and Z.~Kira, ``Coda-prompt: Continual decomposed
  attention-based prompting for rehearsal-free continual learning,'' in
  \emph{Proceedings of the IEEE/CVF Conference on Computer Vision and Pattern
  Recognition}, 2023, pp. 11\,909--11\,919.

\bibitem{103}
Z.~Wang, Z.~Zhang, S.~Ebrahimi, R.~Sun, H.~Zhang, C.-Y. Lee, X.~Ren, G.~Su,
  V.~Perot, J.~Dy \emph{et~al.}, ``Dualprompt: Complementary prompting for
  rehearsal-free continual learning,'' in \emph{European Conference on Computer
  Vision}.\hskip 1em plus 0.5em minus 0.4em\relax Springer, 2022, pp. 631--648.

\bibitem{104}
Y.~Wang, Z.~Huang, and X.~Hong, ``S-prompts learning with pre-trained
  transformers: An occam’s razor for domain incremental learning,''
  \emph{Advances in Neural Information Processing Systems}, vol.~35, pp.
  5682--5695, 2022.

\bibitem{105}
R.~Ji, K.~Tan, X.~Wang, C.~Pan, and L.~Xin, ``Passnet: A spatial–spectral
  feature extraction network with patch attention module for hyperspectral
  image classification,'' \emph{IEEE Geoscience and Remote Sensing Letters},
  vol.~20, pp. 1--5, 2023.

\bibitem{26}
M.~Imani, ``{RX} anomaly detector with rectified background,'' \emph{IEEE
  Geoscience and Remote Sensing Letters}, vol.~14, no.~8, pp. 1313--1317, 2017.

\bibitem{26-1}
H.~Kwon and N.~M. Nasrabadi, ``Kernel {RX}-algorithm: A nonlinear anomaly
  detector for hyperspectral imagery,'' \emph{IEEE transactions on Geoscience
  and Remote Sensing}, vol.~43, no.~2, pp. 388--397, 2005.

\bibitem{26-3}
S.~Matteoli, T.~Veracini, M.~Diani, and G.~Corsini, ``A locally adaptive
  background density estimator: An evolution for {RX}-based anomaly
  detectors,'' \emph{IEEE Geoscience and Remote Sensing Letters}, vol.~11,
  no.~1, pp. 323--327, 2014.

\bibitem{27}
L.~Zhang, J.~Ma, B.~Cheng, and F.~Lin, ``Fractional fourier transform-based
  tensor {RX} for hyperspectral anomaly detection,'' \emph{Remote Sensing},
  vol.~14, no.~3, p. 797, 2022.

\bibitem{28}
F.~He, S.~Yan, Y.~Ding, Z.~Sun, J.~Zhao, H.~Hu, and Y.~Zhu, ``Recursive {RX}
  with extended multi-attribute profiles for hyperspectral anomaly detection,''
  \emph{Remote Sensing}, vol.~15, no.~3, 2023.

\bibitem{32}
Y.~Ma, S.~Cai, and J.~Zhou, ``Adaptive reference-related graph embedding for
  hyperspectral anomaly detection,'' \emph{IEEE Transactions on Geoscience and
  Remote Sensing}, vol.~61, pp. 1--14, 2023.

\bibitem{37}
W.~Li and Q.~Du, ``Collaborative representation for hyperspectral anomaly
  detection,'' \emph{IEEE Transactions on Geoscience and Remote Sensing},
  vol.~53, no.~3, pp. 1463--1474, 2015.

\bibitem{38}
Y.~Xu, Z.~Wu, J.~Li, A.~Plaza, and Z.~Wei, ``Anomaly detection in hyperspectral
  images based on low-rank and sparse representation,'' \emph{IEEE Transactions
  on Geoscience and Remote Sensing}, vol.~54, no.~4, pp. 1990--2000, 2016.

\bibitem{39}
C.-I. Chang, ``Effective anomaly space for hyperspectral anomaly detection,''
  \emph{IEEE Transactions on Geoscience and Remote Sensing}, vol.~60, pp.
  1--24, 2022.

\bibitem{40}
L.~Gao, X.~Sun, X.~Sun, L.~Zhuang, Q.~Du, and B.~Zhang, ``Hyperspectral anomaly
  detection based on chessboard topology,'' \emph{IEEE Transactions on
  Geoscience and Remote Sensing}, vol.~61, pp. 1--16, 2023.

\bibitem{41}
M.~Wang, Q.~Wang, D.~Hong, S.~K. Roy, and J.~Chanussot, ``Learning tensor
  low-rank representation for hyperspectral anomaly detection,'' \emph{IEEE
  Transactions on Cybernetics}, vol.~53, no.~1, pp. 679--691, 2023.

\bibitem{42}
C.~Zhao, C.~Li, S.~Feng, and X.~Jia, ``Enhanced total variation regularized
  representation model with endmember background dictionary for hyperspectral
  anomaly detection,'' \emph{IEEE Transactions on Geoscience and Remote
  Sensing}, vol.~60, pp. 1--12, 2022.

\bibitem{43}
S.~Mohla, S.~Pande, B.~Banerjee, and S.~Chaudhuri, ``{FusAtNet}: Dual attention
  based spectrospatial multimodal fusion network for hyperspectral and lidar
  classification,'' in \emph{Proceedings of the IEEE/CVF Conference on Computer
  Vision and Pattern Recognition (CVPR) Workshops}, June 2020, pp. 92--93.

\bibitem{44}
L.~Gao, J.~Li, K.~Zheng, and X.~Jia, ``Enhanced autoencoders with
  attention-embedded degradation learning for unsupervised hyperspectral image
  super-resolution,'' \emph{IEEE Transactions on Geoscience and Remote
  Sensing}, vol.~61, pp. 1--17, 2023.

\bibitem{45}
Z.~Wu, M.~E. Paoletti, H.~Su, X.~Tao, L.~Han, J.~M. Haut, and A.~Plaza,
  ``Background-guided deformable convolutional autoencoder for hyperspectral
  anomaly detection,'' \emph{IEEE Transactions on Geoscience and Remote
  Sensing}, vol.~61, pp. 1--16, 2023.

\bibitem{46}
S.~Wang, X.~Wang, L.~Zhang, and Y.~Zhong, ``{Auto-AD}: Autonomous hyperspectral
  anomaly detection network based on fully convolutional autoencoder,''
  \emph{IEEE Transactions on Geoscience and Remote Sensing}, vol.~60, pp.
  1--14, 2022.

\bibitem{47}
S.~Arisoy, N.~M. Nasrabadi, and K.~Kayabol, ``{GAN}-based hyperspectral anomaly
  detection,'' in \emph{2020 28th European Signal Processing Conference
  (EUSIPCO)}, 2021, pp. 1891--1895.

\bibitem{wang2021dual}
J.~Wang, S.~Guo, R.~Huang, L.~Li, X.~Zhang, and L.~Jiao, ``Dual-channel capsule
  generation adversarial network for hyperspectral image classification,''
  \emph{IEEE Transactions on Geoscience and Remote Sensing}, vol.~60, pp.
  1--16, 2021.

\bibitem{48}
K.~Jiang, W.~Xie, Y.~Li, J.~Lei, G.~He, and Q.~Du, ``Semisupervised spectral
  learning with generative adversarial network for hyperspectral anomaly
  detection,'' \emph{IEEE Transactions on Geoscience and Remote Sensing},
  vol.~58, no.~7, pp. 5224--5236, 2020.

\bibitem{49}
Z.~He, D.~He, M.~Xiao, A.~Lou, and G.~Lai, ``Convolutional transformer-inspired
  autoencoder for hyperspectral anomaly detection,'' \emph{IEEE Geoscience and
  Remote Sensing Letters}, vol.~20, pp. 1--5, 2023.

\bibitem{106}
\BIBentryALTinterwordspacing
Z.~Wang, X.~Wang, K.~Tan, B.~Han, J.~Ding, and Z.~Liu, ``Hyperspectral anomaly
  detection based on variational background inference and generative
  adversarial network,'' \emph{Pattern Recognition}, vol. 143, p. 109795, 2023.
  [Online]. Available:
  \url{https://www.sciencedirect.com/science/article/pii/S0031320323004934}
\BIBentrySTDinterwordspacing

\bibitem{83}
C.~V. Nguyen, Y.~Li, T.~D. Bui, and R.~E. Turner, ``Variational continual
  learning,'' \emph{arXiv preprint arXiv:1710.10628}, 2017.

\bibitem{yuhas1992discrimination}
R.~H. Yuhas, A.~F. Goetz, and J.~W. Boardman, ``Discrimination among semi-arid
  landscape endmembers using the spectral angle mapper ({SAM}) algorithm,'' in
  \emph{JPL, Summaries of the Third Annual JPL Airborne Geoscience Workshop.
  Volume 1: AVIRIS Workshop}, 1992.

\bibitem{80}
G.~E. Hinton and R.~Zemel, ``Autoencoders, minimum description length and
  helmholtz free energy,'' \emph{Advances in neural information processing
  systems}, vol.~6, 1993.

\bibitem{81}
Q.~Jin, Y.~Ma, X.~Mei, and J.~Ma, ``Tanet: An unsupervised two-stream
  autoencoder network for hyperspectral unmixing,'' \emph{IEEE Transactions on
  Geoscience and Remote Sensing}, vol.~60, pp. 1--15, 2021.

\bibitem{82}
Y.~Li, Y.~Shi, K.~Wang, B.~Xi, J.~Li, and P.~Gamba, ``Target detection with
  unconstrained linear mixture model and hierarchical denoising autoencoder in
  hyperspectral imagery,'' \emph{IEEE Transactions on Image Processing},
  vol.~31, pp. 1418--1432, 2022.

\bibitem{EM}
A.~P. Dempster, N.~M. Laird, and D.~B. Rubin, ``Maximum likelihood from
  incomplete data via the em algorithm,'' \emph{Journal of the royal
  statistical society: series B (methodological)}, vol.~39, no.~1, pp. 1--22,
  1977.

\bibitem{52}
X.~Kang, X.~Zhang, S.~Li, K.~Li, J.~Li, and J.~A. Benediktsson, ``Hyperspectral
  anomaly detection with attribute and edge-preserving filters,'' \emph{IEEE
  Transactions on Geoscience and Remote Sensing}, vol.~55, no.~10, pp.
  5600--5611, 2017.

\bibitem{91}
Z.~Li, Y.~Wang, C.~Xiao, Q.~Ling, Z.~Lin, and W.~An, ``You only train once:
  Learning a general anomaly enhancement network with random masks for
  hyperspectral anomaly detection,'' \emph{IEEE Transactions on Geoscience and
  Remote Sensing}, vol.~61, pp. 1--18, 2023.

\bibitem{53}
R.~Aljundi, F.~Babiloni, M.~Elhoseiny, M.~Rohrbach, and T.~Tuytelaars, ``Memory
  aware synapses: Learning what (not) to forget,'' in \emph{Proceedings of the
  European Conference on Computer Vision (ECCV)}, September 2018, pp. 139--154.

\bibitem{54}
J.~Kirkpatrick, R.~Pascanu, N.~Rabinowitz, J.~Veness, G.~Desjardins, A.~A.
  Rusu, K.~Milan, J.~Quan, T.~Ramalho, A.~Grabska-Barwinska, D.~Hassabis,
  C.~Clopath, D.~Kumaran, and R.~Hadsell, ``Overcoming catastrophic forgetting
  in neural networks,'' \emph{Proceedings of the National Academy of Sciences},
  vol. 114, no.~13, pp. 3521--3526, 2017.

\bibitem{55}
G.~Zeng, Y.~Chen, B.~Cui, and S.~Yu, ``Continual learning of context-dependent
  processing in neural networks,'' \emph{Nature Machine Intelligence}, vol.~1,
  no.~8, pp. 364--372, 2019.

\bibitem{60}
J.~Wang, S.~Guo, Z.~Hua, R.~Huang, J.~Hu, and M.~Gong, ``{CL-CaGAN}: Capsule
  differential adversarial continual learning for cross-domain hyperspectral
  anomaly detection,'' \emph{IEEE Transactions on Geoscience and Remote
  Sensing}, vol.~62, pp. 1--15, 2024.

\bibitem{zweig1993receiver}
M.~H. Zweig and G.~Campbell, ``Receiver-operating characteristic ({ROC}) plots:
  a fundamental evaluation tool in clinical medicine,'' \emph{Clinical
  chemistry}, vol.~39, no.~4, pp. 561--577, 1993.

\bibitem{ferri2011coherent}
C.~Ferri, J.~Hern{\'a}ndez-Orallo, and P.~A. Flach, ``A coherent interpretation
  of auc as a measure of aggregated classification performance,'' in
  \emph{Proceedings of the 28th International Conference on Machine Learning
  (ICML-11)}, 2011, pp. 657--664.

\bibitem{chen2021component}
S.~Chen, C.-I. Chang, and X.~Li, ``Component decomposition analysis for
  hyperspectral anomaly detection,'' \emph{IEEE Transactions on Geoscience and
  Remote Sensing}, vol.~60, pp. 1--22, 2021.

\end{thebibliography}




-








\vspace{-18mm}
\begin{IEEEbiography}[{\includegraphics[width=0.8in,height=1in,clip]{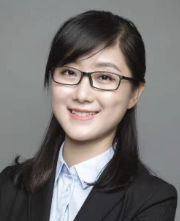}}]{Jianing Wang}
(Member, IEEE) received the B.S. and M.S. degrees in circuit and system from Lanzhou University, Lanzhou, China, in 2005 and 2008, respectively, and the Ph.D. degree from Xidian University, Xi’an, China, in 2016. 

She worked with China Aerospace Science and Technology Corporation, Xi’an. She is an Associate Professor and a Member of the Key Laboratory
of Intelligent Perception and Image Understanding, Ministry of Education of China, School of Computer Science and Technology, Xidian University, Xi’an. Her research interests include image processing, machine learning, and artificial intelligent algorithm and applications. Her research directions include big data processing, embedded algorithms for intelligent model compression methods, pattern recognition and artificial intelligence, video data processing, analysis and content understand and continual learning.
\end{IEEEbiography}
\vspace{-18mm}
\begin{IEEEbiography}[{\includegraphics[width=0.75in,height=1.1in,clip]{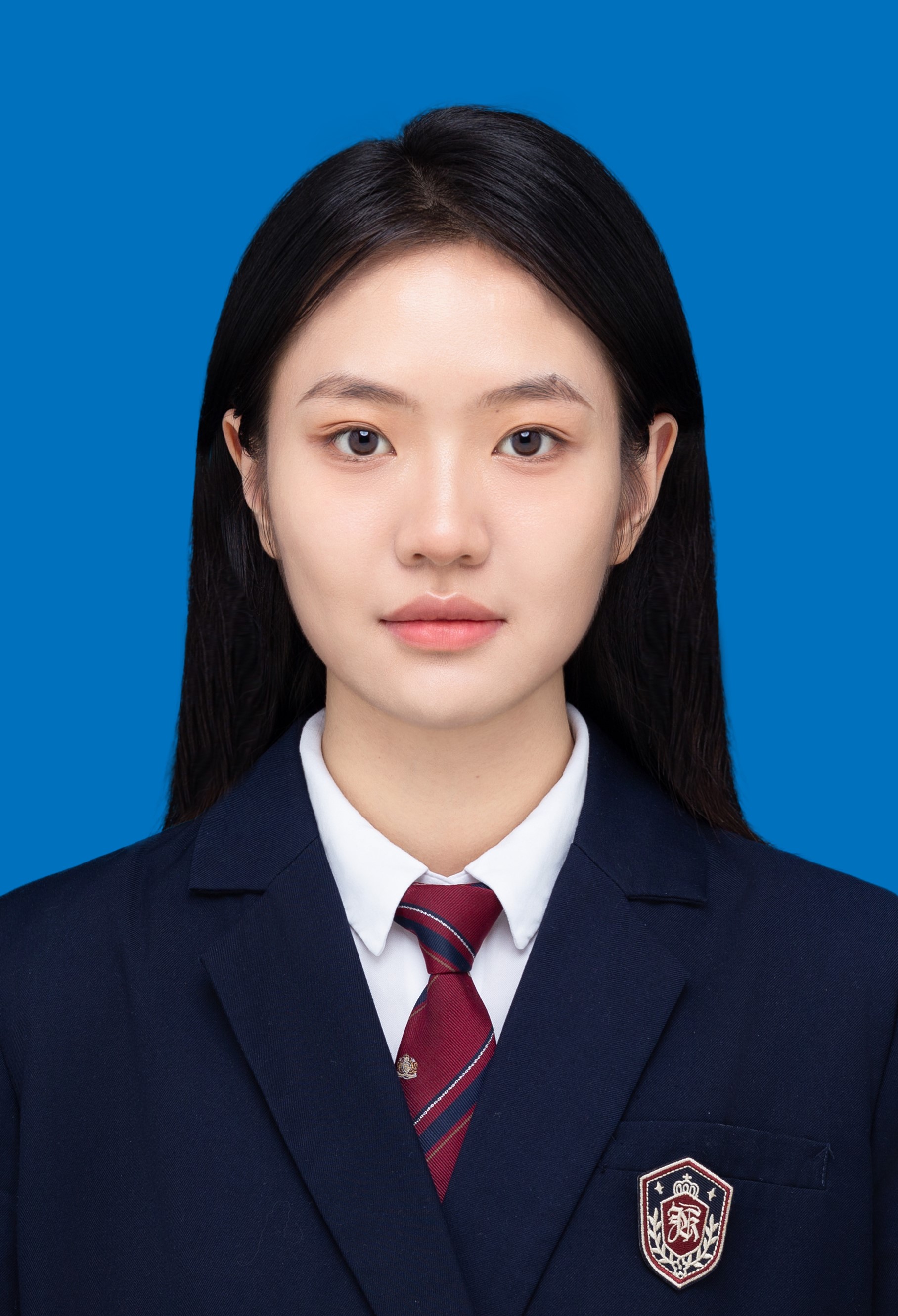}}]{Zheng Hua} received the Bachelor’s degree from Digital Media Technology in 2022, Xi'an Engineering University, Xi‘an, China, where she is pursuing a Master's degree in Computer Science and Technology from 2022, Xidian University, Xi’an, China.
 
Her research interests include continual learning, incremental learning, hyperspectral anomaly detection and image processing.
\end{IEEEbiography}

 \vspace{-18mm}
\begin{IEEEbiography}[{\includegraphics[width=0.8in,height=1in,clip]{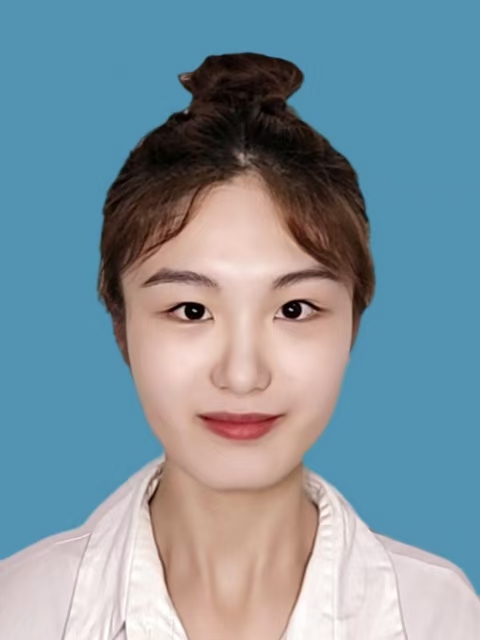}}]{Wan Zhang} received the Bachelor's degree from Intelligent Science and Technology in 2024, Jiangsu Normal University, Xuzhou, China, where she is pursuing a Master's degree in Computer Science and Technology from Xidian University in 2024, Xi'an, China.

Her research interests include continual learning, prompt learning, hyperspectral large model and image processing.
\end{IEEEbiography}

\vspace{-18mm}
\begin{IEEEbiography}[{\includegraphics[width=0.8in,height=1in,clip,keepaspectratio]{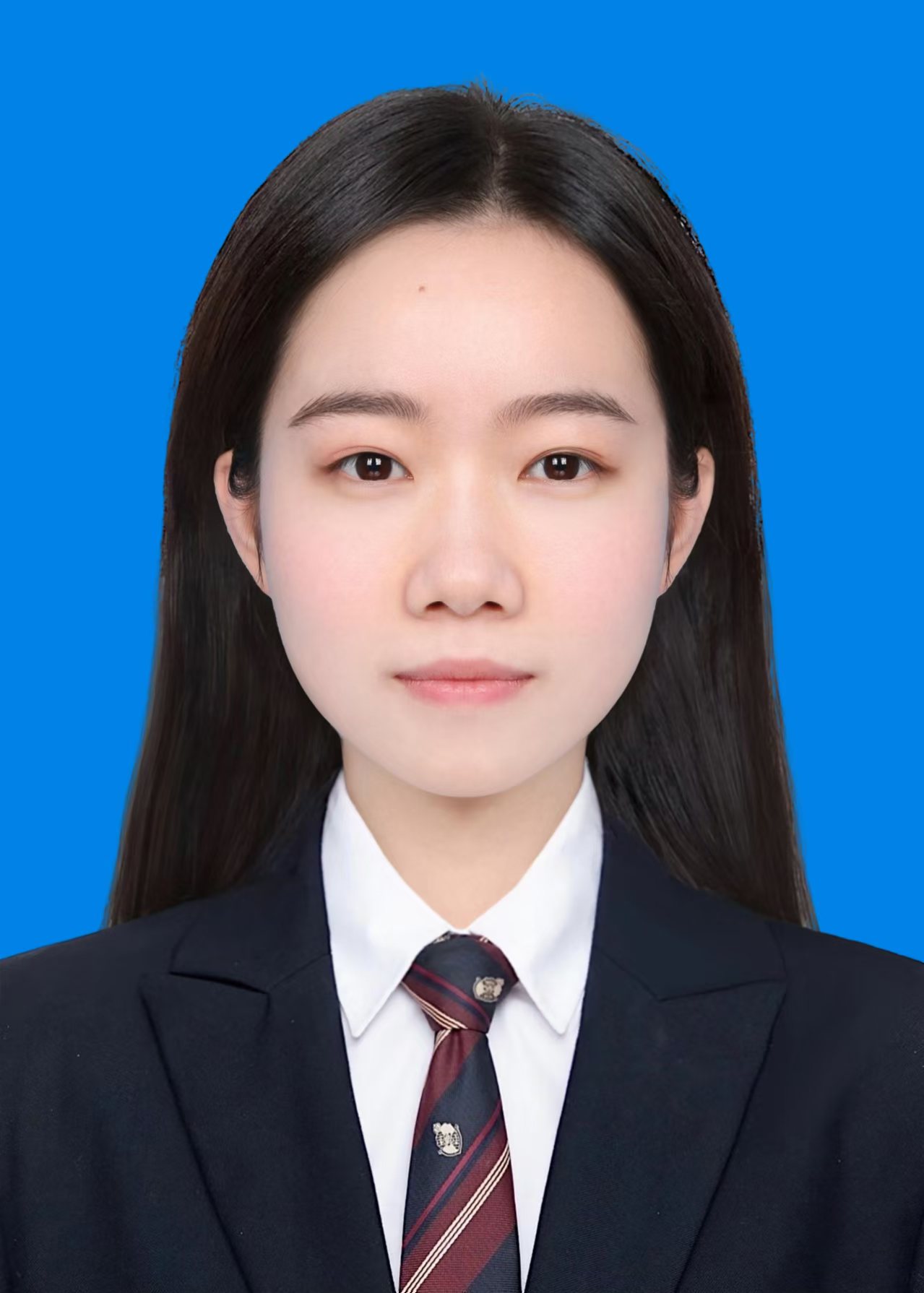}}]{ShengJia Hao} received the Bachelor’s degree from Digital Media Technology in 2022, Xi'an University of Technology, Xi’an, China, where she is pursuing a Master's degree in Computer Science and Technology at Xidian University from 2022 in Xi'an, China.

Her research interests include point cloud classification, point cloud segmentation, 3D object detection, 3D vision, and multi-modal learning.  
\end{IEEEbiography}

\vspace{-18mm}
\begin{IEEEbiography}[{\includegraphics[width=0.8in,height=1in,clip,keepaspectratio]{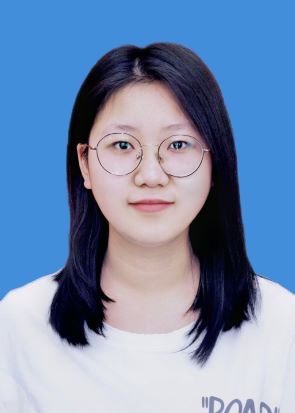}}]{YuQiong Yao} received the Bachelor’s degree from Data Science and Big Data Technology in 2022, Shanxi University, Xi’an, China. She is pursuing a Master's degree in Computer Science and Technology at Xidian University from 2022 in Xi'an, China.

Her research interests include video action recognition, video action detection and prompt learning.
\end{IEEEbiography}
\begin{IEEEbiography}[{\includegraphics[width=0.8in,height=1in,clip,keepaspectratio]{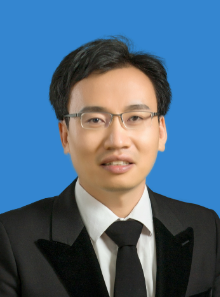}}]{Maoguo Gong} (Fellow, IEEE) received the B. Eng degree and Ph.D. degree from Xidian University. Since 2006, he has been a teacher of Xidian University. He was promoted to associate professor and full professor in 2008 and 2010, respectively, both with exceptive admission.
	
Gong’s research interests are broadly in the area of computational intelligence, with applications to optimization, learning, data mining and image understanding. He has published over one hundred papers in journals and conferences, and holds over twenty granted patents as the first inventor. He is leading or has completed over twenty projects as the Principle Investigator, funded by the National Natural Science Foundation of China, the National Key Research and Development Program of China, and others. He was the recipient of the prestigious National Program for Support of the Leading Innovative Talents from the Central Organization Department of China, the Leading Innovative Talent in the Science and Technology from the Ministry of Science and Technology of China, the Excellent Young Scientist Foundation from the National Natural Science Foundation of China, the New Century Excellent Talent from the Ministry of Education of China, and the National Natural Science Award of China.
	
He is the Executive Committee Member of Chinese Association for Artificial Intelligence, Senior Member of IEEE and Chinese Computer Federation, Associate Editor or Editorial Board Member for over five journals including the IEEE Transactions on Evolutionary Computation and the IEEE Transactions on Neural Networks and Learning Systems.
 \end{IEEEbiography}

\end{document}